\documentclass{article}

\usepackage{natbib}
\setcitestyle{numbers,square}

    \usepackage[final]{neurips_2024}

\usepackage[utf8]{inputenc} 
\usepackage[T1]{fontenc}    
\usepackage{hyperref}       
\usepackage{url}            
\usepackage{booktabs}       
\usepackage{amsfonts}       
\usepackage{nicefrac}       
\usepackage{microtype}      
\usepackage[table]{xcolor}        

\usepackage{amssymb}
\usepackage{amsmath}
\usepackage{amsthm}
\usepackage{graphicx}
\usepackage{subfigure}
\usepackage{wrapfig}
\usepackage{float}

\usepackage{graphicx} 

\usepackage{algorithm}
\usepackage{algorithmic}

\usepackage{multirow}
\usepackage{enumerate}

\usepackage{makecell}

\newtheorem{theorem}{Theorem}[section]

\newtheorem{lemma}[theorem]{Lemma}

\newtheorem{definition}[theorem]{Definition}

\title{Spiking Graph Neural Network on Riemannian Manifolds}

\author{%
      Li Sun$^1$, Zhenhao Huang$^1$, Qiqi Wan$^1$, Hao Peng$^2$, Philip S. Yu$^3$ \\
  $^1$North China Electric Power University, Beijing, China\\
    $^2$Beihang University, Beijing, China\\
      $^3$University of Illinois at Chicago, IL, USA\\
      \texttt{ccesunli@ncepu.edu.cn, penghao@buaa.edu.cn, psyu@uic.edu} \\
}
\begin{document}
\maketitle

\begin{abstract}
Graph neural networks (GNNs) have become the dominant solution for learning on graphs, the typical non-Euclidean structures.
Conventional GNNs, constructed with the Artificial Neuron Network (ANN), have achieved impressive performance at the cost of high computation and energy consumption.
In parallel, spiking GNNs with brain-like spiking neurons are drawing increasing research attention owing to the energy efficiency.
So far, existing spiking GNNs consider graphs in Euclidean space, ignoring the structural geometry, and suffer from the high latency issue due to Back-Propagation-Through-Time (BPTT) with the surrogate gradient.
In light of the aforementioned issues, \emph{we are devoted to exploring spiking GNN on Riemannian manifolds}, and present a Manifold-valued Spiking GNN (\texttt{MSG}).
In particular, we design a new spiking neuron on geodesically complete manifolds with the diffeomorphism, so that BPTT regarding the spikes is replaced by the proposed differentiation via manifold.
Theoretically, we show that \texttt{MSG}  approximates a solver of the manifold ordinary differential equation.
Extensive experiments on common graphs show the proposed \texttt{MSG} achieves superior performance to previous spiking GNNs and energy efficiency to conventional GNNs.
\end{abstract}


 \vspace{-0.15in}
\section{Introduction}
 \vspace{-0.1in}

Graphs are the ubiquitous, non-Euclidean structures that describe the relationship among objects. 
Graph neural networks (GNNs), constructed with the floating-point Artificial Neuron Network (ANN), have achieved state-of-the-art accuracy for learning on graphs \cite{hamilton2017inductive,velickovic2018graph,chami2019hyperbolic,xiong2022pseudo}.
However, they raise the concerns about computation and energy consumption, particularly when dealing with real-world graphs of considerable scale \cite{zhu2022spiking,li2023graph}.
In contrast, Spiking Neuron Networks (SNNs), inspired by the biological mechanism of brains, utilize neurons that communicate using sparse and discrete spikes, showcasing their superiority in energy efficiency \cite{maass1997networks,brette2007simulation}.
Attempting to bring the best of both worlds, \textbf{spiking GNNs} are drawing increasing research attention.

In the literature of spiking GNNs, recent efforts have been made to design different architectures with spiking neurons, e.g., graph convolution \cite{zhu2022spiking}, attention mechanism \cite{wang2022spiking}, variational autoencoder \cite{yang2022spiking} and continuous GNN \cite{yin2024continuous}. 
While achieving encouraging results, existing spiking GNNs still face several fundamental issues:
\textbf{(1) Representation Space.} 
Spiking GNNs consider the graph in Euclidean space, ignoring the inherent geometry of graph structures.
Unlike the Euclidean structures (e.g., pixel matrix and grid structures), graphs cannot be embedded in Euclidean space with bounded distortion \cite{sarkar2012low}.
Instead, \emph{Riemannian manifolds} have been shown as the promising spaces to model graphs in recent years \cite{chami2019hyperbolic,bachmann2020constanta,xiong2022pseudo} (e.g., hyperbolic space, a type of Riemannian manifolds, is well aligned with the graphs dominated by hierarchical structures).
However, none of the existing works study SNN on Riemannian manifolds, to the best of our knowledge. It is thus an interesting and urgent problem to consider how to endow the spiking GNN with a Riemannian manifold.
\textbf{(2) Training Algorithm.} 
Training spiking GNN is challenging, since the spikes are non-differentiable. 
Existing studies consider the spiking GNN as a recurrent neural network and apply Backward-Passing-Through-Time (BPTT) with the surrogate gradient  \cite{zhu2022spiking,wang2022spiking,yang2022spiking,yin2024continuous}.
They recurrently compute the backward gradient at each time step, and thus suffer from the high latency issue \cite{wu2021training,meng2022training,wu2023tandem, li2023graph} especially when the spike trains are long.



\begin{wrapfigure}{r}{0.6\textwidth}
\centering
 \vspace{-0.1in}
\includegraphics[width=1\linewidth]{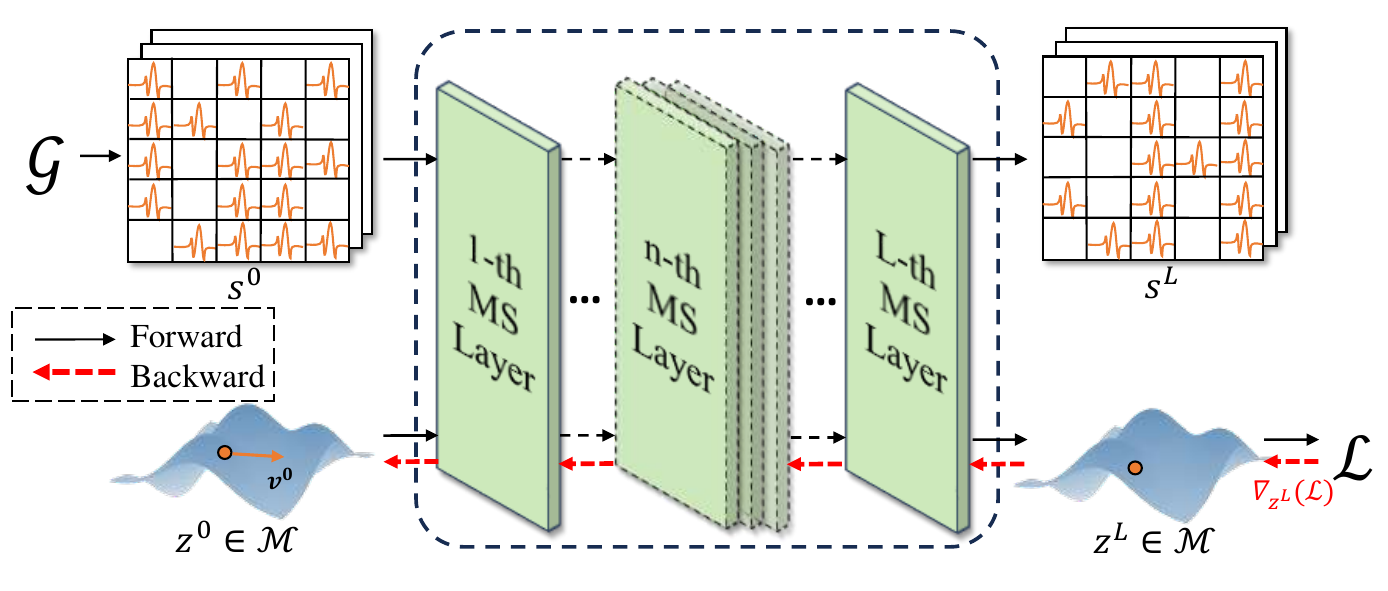}
 \vspace{-0.2in}
\caption{\texttt{MSG} conducts parallel forwarding and enables a new training algorithm alleviating the high latency issue.}
\vspace{-0.1in}
\label{fig. overall-archi}
\end{wrapfigure}


\vspace{-0.1in}
\paragraph{Present work.} Deviating from previous spiking GNNs in Euclidean space, 
in this paper, we open a new direction to explore spiking GNNs on Riemannian manifolds, and propose a novel Manifold-valued Spiking GNN (\texttt{MSG}) sketched in Fig. \ref{fig. overall-archi}.
It is not realistic to place spike trains in a manifold such as hyperbolic or hyperspherical space, given the fact that spike trains cannot align with the defining domain.
Instead, we design a Manifold Spiking Layer that conducts parallel forwarding of spike trains and manifold representations.
Specifically, 
we first incorporate the structural information into spike trains by graph convolution.
Then,  a new \emph{manifold spiking neuron} is proposed to emit spike trains and relate them to manifold representations with \emph{diffeomorphism},
where the spike train generates a momentum that forwards manifold representation along the geodesic.
Instead of applying BPTT in spike domain, the proposed neuron provides us with an alternative of \emph{Differentiation via Manifold} (\emph{DvM}).
(The {\color{red}{red dashed line}} in Fig. \ref{fig. overall-archi}.)
Yet, differentiation in Riemannian manifold is nontrivial.
We leverage the properties of pullback and derive the closed-form backward gradient (Theorem 4.1).
\emph{DvM} enables the recurrence-free gradient backpropagation, which no longer needs to perform recurrent computation of time steps as in BPTT.
Theoretically, \texttt{MSG} is essentially related to manifold Ordinary Differential Equation (ODE).
Each layer creates a \emph{chart} of the manifold, and \texttt{MSG} approximates the dynamic chart solver \cite{lou2020neural} of manifold ODE (Theorem 5.2).



\vspace{-0.1in}
\paragraph{Contributions.} Overall, the key contributions are summarized as follows:
(1) To the best of our knowledge, we propose the first spiking neural network on Riemannian manifolds (\texttt{MSG})\footnote{Codes are available at \url{https://github.com/ZhenhHuang/MSG}}, and show its connection to manifold ODE theoretically.
(2) We design a new training algorithm of differentiation via manifold, which avoids the high latency of BPTT methods.
(3) Extensive experiments  show the superior effectiveness and energy efficiency of the proposed \texttt{MSG}.

\vspace{-0.12in}

\section{Related Work}
\vspace{-0.03in}

We briefly overview the ANN-based GNNs (i.e., the conventional, floating-point GNNs living in either Euclidean space or Riemannian manifolds) and SNN-based GNNs (i.e., spiking GNNs).

\vspace{-0.1in}

\paragraph{ANN-based GNNs (Euclidean and Riemannian).}


The majority of GNNs are built with floating-point ANN, conducting message passing on the graphs \cite{kipf2016semi,velickovic2018graph,wu2019simplifying}.
The Euclidean space has been the workhorse for graph representation learning for decades, and the popular GCN \cite{kipf2016semi}, GAT \cite{velickovic2018graph} and SGC \cite{wu2019simplifying} are also designed in the Euclidean space.
In recent years, Riemannian manifolds have emerged as an exciting alternative considering the geometry of graph structures \cite{feng2019hypergrapha,guo2021hierarchicala}. 
Among Riemannian manifolds, hyperbolic space is recognized for its alignment with the graphs of hierarchical structures, and a series of hyperbolic GNNs (e.g., HGNN \cite{liu2019hyperbolica}, HGCN \cite{chami2019hyperbolic}) show superior performance to their Euclidean versions.
Beyond hyperbolic space, hyperspherical space is well suited for cyclical structures \cite{coors2018spherenet}, and recent studies further investigate the constant curvature spaces \cite{bachmann2020constanta}, product spaces \cite{gu2018learning, zhang2021switch,aaai22SelfMix,SunL24AAAI}, quotient spaces \cite{law2021ultrahyperbolic}, SPD manifolds \cite{gao2020learning, dong2017deep}, etc. 
Riemannian manifolds achieve remarkable success in graph clustering \cite{SunL24ICML,SunL24WWW}, structural learning \cite{SunL23ICDM}, graph dynamics \cite{SunL23AAAI,HVGNN,SunL22CIKM} and information diffusion \cite{SunL24SIGIR}, but have rarely been touched yet in the SNN counterpart.

\vspace{-0.1in}
\paragraph{Spiking Neural Networks (SNNs) \& Spiking GNNs.}

Mimicking the biological neural networks, SNNs \cite{maass1997networks,brette2007simulation} utilize the spiking neuron to process spike trains, and offer the advantage of energy efficiency.
Despite the wide application of SNN in computer vision \cite{cao2015spiking,cao2024spiking}, SNNs are still at an early stage in the graph domain. 
The basic idea of spiking GNNs is adapting ANN-based GNNs to the SNN framework  by substituting the activation functions with spiking neurons. 
Pioneering works study the graph convolution \cite{xu2021exploiting,zhu2022spiking}, and efforts have also been made to the graph attention \cite{wang2022spiking}, variational graph autoencoder \cite{yang2022spiking}, graph differential equations \cite{poli2019graph}, etc.
SpikeGCL \cite{li2023graph} is a recent endeavor to conduct graph contrastive learning with SNN.
In parallel, spiking GNNs are extended to model the dynamic graphs \cite{li2023scaling,zhao2024dynamic,yin2024dynamic}. We focus on the static graph in this work.
In both dynamic and static cases, previous spiking GNNs are trained with the surrogate gradient, leading to high latency, and consider the graphs in the Euclidean space.

\vspace{-0.12in}

\section{Preliminaries}
 \vspace{-0.1in}
Different from aforementioned spiking GNNs, we study the spiking GNN on Riemannian manifolds.
Thus, we formally introduce the basic concepts of Riemannian geometry and SNN.
Throughout this paper, the lowercase boldfaced $\boldsymbol x$ and uppercase $\mathbf X$ denote vector and matrix, respectively. Important notations are summarized in Appendix \ref{append. notation}.

\vspace{-0.12in}

\paragraph{Riemannian Geometry \& Riemannian Manifold.}

Riemannian geometry provides elegant framework to study structures and manifolds.
A Riemannian manifold is described as a smooth and real manifold $\mathcal{M}$ endowed with a Riemannian metric.
Each point $\boldsymbol x$ in the manifold is associated with the \emph{tangent space} $T_{\boldsymbol x}\mathcal M$ that ``looks Euclidean'',  and the Riemannian metric is given by the inner product in the tangent space, so that geometric properties (e.g, angle, length) can be defined.
A \emph{geodesic} between two points on the manifold is the smooth path connecting them with the minimal length. 
There exist three types of isotropic manifold, namely, the \emph{Constant Curvature Space} (CCS): hyperbolic space $\mathbb H$, hyperspherical space $\mathbb S$ and the special case of Euclidean space with ``flat'' geometry $\mathbb E$ .


\vspace{-0.12in}

\paragraph{Graph \& Riemannian Graph Representation Learning.}
A graph $\mathcal{G} = (\mathcal{V},\mathcal{E}, \mathbf{F}, \mathbf{A})$ is defined on the node set $\mathcal{V}$  and edge set $\mathcal{E}\subset \mathcal{V} \times \mathcal{V}$, and $\mathbf A\in \mathbb R^{|\mathcal{V}|\times |\mathcal{V}|}$ is the adjacency matrix describing the structure information.
Each node $v_i$ is associated with a feature  $\boldsymbol f_i$, and node features are summarized in $\mathbf{F} \in \mathbb{R}^{|\mathcal{V}|\times d}$.
In this paper, we resolve the problem of \emph{Riemannian Graph Representation Learning} with SNN. 
Specifically, we seek a graph encoder  $\mathcal{F}_\theta: v \mapsto \boldsymbol z $ where $\boldsymbol z\in \mathcal M$ is a point on the manifold, instead of Euclidean space, and $\mathcal{F}_\theta$ is defined with an energy-efficient SNN.

\vspace{-0.12in}

\paragraph{Spiking Neural Network.}
SNNs are constructed by \emph{spiking neurons} that communicate with each other by spike trains.
Concretely, a spiking neuron is conceptualized as ``a capacitor of the membrane potential'', and processes the spike trains by the following $3$ phases \cite{salinas2002integrate}.
First, the incoming current $I[t]$ is accumulated in the capacitor, leading to the potential buildup (\emph{integrate}). 
When the membrane potential $V[t]$ reaches or exceeds a specific threshold $V_{th}$, the neuron emits a spike (\emph{fire}). 
After that, the membrane potential is reset to the resting potential $V_{reset}$ (\emph{reset}). 
There are two popular spiking neurons: IF model and LIF model \cite{burkitt2006review}. 
In particular, the three  phases of IF model are formalized as
\vspace{-0.01in}
\begin{align}
     \operatorname{Integrate}: \ \ V[t] &=g(V[t-1], I[t]) =  V[t-1] + I[t]   \label{eq.IF-integrate}          \\
     \operatorname{Fire}: \ \ S[t] &=H(V[t]-V_{th})               \\
     \operatorname{Reset}: \ \ V[t] &=\left \{
        \begin{aligned}
        & (1-S[t])V[t] + S[t]V_{rest}, &\textit{Fixed-reset}, \\
        & V[t]-V_{th}S[t], &\textit{Subtraction-reset}. 
        \end{aligned}
        \right. 
\end{align}
where the incoming current $I[t]$ is related to the input spike train, and  $V_{reset}$ is lower than $V_{th}$. $t$ denotes the time index of the spike. 
The Heaviside function $H(\cdot)$ is non-differentiable, $H(x)=1$ if $x \ge 0$, and $0$ otherwise.
There are two options for reset, and fixed-reset is adopted in this paper.
Overall, an IF model is given as $S[t] =\operatorname{IFModel}(I[t]) $, and the only difference between IF model and LIF model lies in the definition of $g(\cdot)$ in Eq. (\ref{eq.IF-integrate}).
In this paper, we are interested in designing a new spiking neuron on Riemannian manifold.

\vspace{-0.12in}

\section{Methodology: Manifold-valued Spiking GNN}
\vspace{-0.1in}

In this section, we present a simple yet effective Manifold-valued Spiking GNN (\texttt{MSG}), which can be applied to any geodesically complete manifolds,
e.g., the Constant Curvature Space (CCS), including hyperbolic space and hyperspherical space, or the product of CCS.
In particular, we design a spiking neuron on Riemannian manifolds (named as \emph{Manifold Spiking Neuron}) that allows for the \emph{differentiation via manifold}.
It provides a new perspective of training spiking GNN, so that we avoid the high latency of typical backward-passing-through-time (BPTT) training.




\vspace{-0.12in}

\subsection{Manifold Spiking Layer}
\vspace{-0.1in}

We elaborate on the sole building block of the proposed model --- Manifold Spiking Layer.
Note that, the spike train or spiking representation in existing spiking GNNs \cite{xu2021exploiting,zhu2022spiking,wang2022spiking,yang2022spiking,li2023graph,li2023scaling,zhao2024dynamic,yin2024dynamic} cannot align with the defining domain of Riemannian manifolds (e.g., hyperbolic space and hyperspherical space), thus posing a fundamental challenge. Our solution is to generate node representation on the manifold (referred to as manifold representation) in parallel, and leverage the notion of Diffeomorphism  to create the alignment between the two domains. We formulate the \emph{parallel forwarding} of spike trains and manifold representations as follows.


\begin{wrapfigure}{l}{0.5\textwidth}
\centering
\includegraphics[width=1\linewidth]{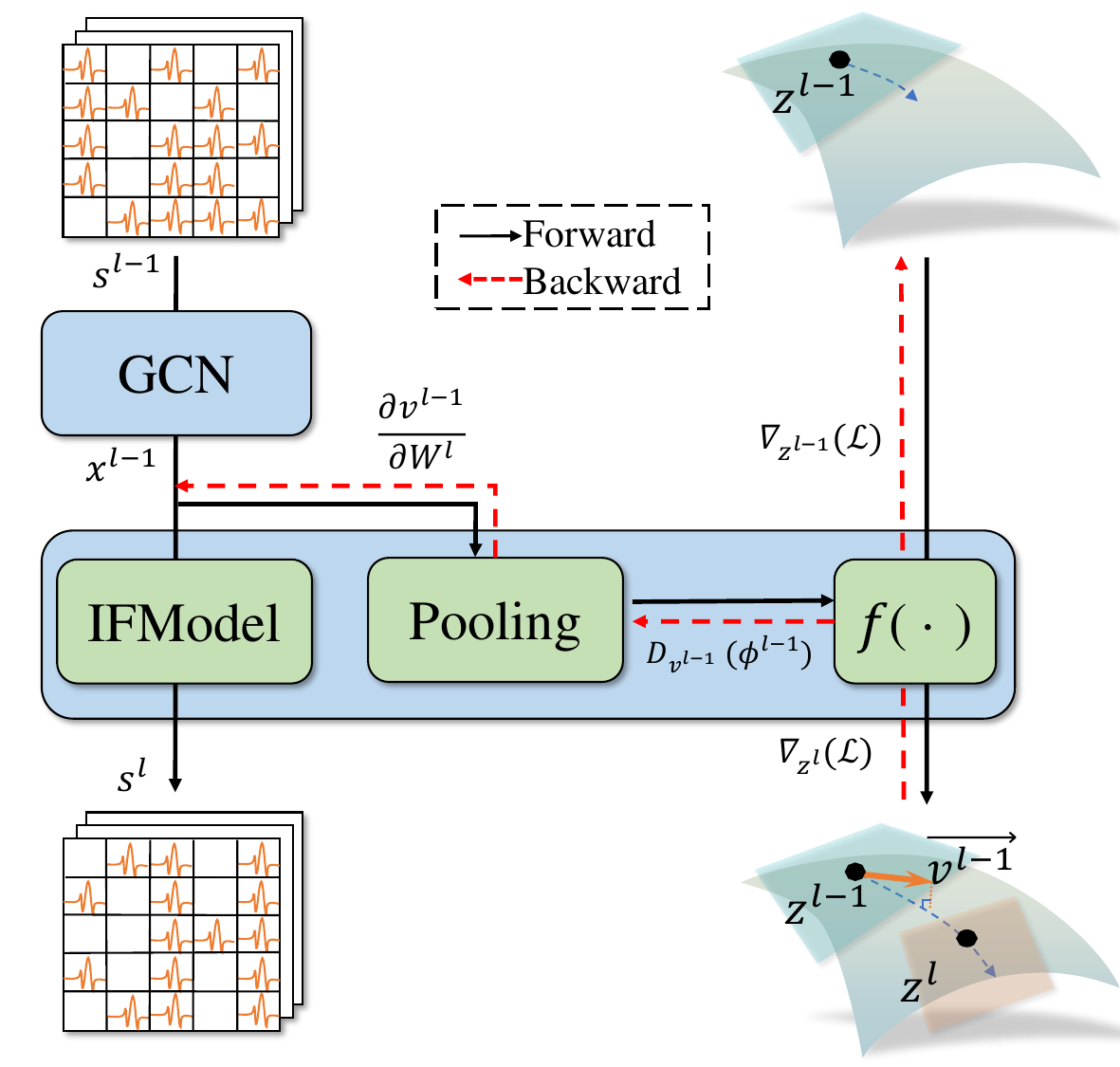}
 \vspace{-0.2in}
\caption{Manifold Spiking Layer. It conducts parallel forwarding of spike trains and manifold representations, and creates an alternative  backward pass ({\color{red}{red dashed line}}). The backward gradient with 
$\frac{\partial \mathbf{v}^{l-1}}{\partial \mathbf{W}^l}$, $ D_{\mathbf{v}^{l-1}}\phi^{l-1}$ and $\nabla_{\mathbf{z}^l} \mathcal{L}$ will be introduced in Sec. \ref{sec:backward}.
}
\vspace{-0.5in}
\label{fig. neuron}
\end{wrapfigure}

\vspace{-0.1in}
\paragraph{Unified Formulation.}
The forward pass of the spiking layer consists of a graph convolution and one proposed manifold spiking neuron.
Without loss of generality, for each node $v_i \in \mathcal G$, the $l-$th spiking layer is formulated as follows,
\begin{align}
     \mathbf{x}_i^{l-1}[t] &= \operatorname{GCN}(\mathbf{s}_i^{l-1}[t];\mathbf{W}^l) , \label{eq. x}\\
     [\mathbf{s}_i^l, \mathbf{z}_i^l ] & = \operatorname{MSNeuron}(\mathbf{x}_i^{(l-1)}, \mathbf{z}_i^{(l-1)}),
\end{align}
where $\mathbf x$ is the incoming current to generate spike trains. $\mathbf s$ and $\mathbf z$ denote the spike trains and manifold representation, respectively.
$\operatorname{GCN}(\cdot)$ is a GNN in the Euclidean space, and $\mathbf{W}^l$ is the learnable parameter in the layer.
Different from the neuron of previous spiking GNNs, 
we design \emph{a novel manifold spiking neuron} ($\operatorname{MSNeuron}$) as shown in Fig. \ref{fig. neuron}.
It emits spike trains and relates them to manifold representations simultaneously, which is formulated as follows,
\begin{align}
\label{eq. input}
\mathbf{s}_i^l &= \operatorname{IFModel}(\{\mathbf{x}_i^{l-1}[t]\}_{t=1,...,T}) \\
\mathbf{v}_i^{l-1} &=\operatorname{Pooling}(\mathbf{x}_i^{l-1}[t])  \\
\mathbf{z}_i^l &= f (\mathbf{z}_i^{l-1}, \epsilon\mathbf{v}_i^{l-1}) \in \mathcal M
    \label{eq. z}
\end{align}
where $t$ is the time step of spike trains.
The IF model can be replaced by LIF model, and we utilize IF model by default for simplicity.
$\operatorname{Pooling}$ is defined as the average pooling of the current $\mathbf{x}$ over $t$, and  
$\mathbf{v}$ is given as the Euclidean vector.
$f$ denotes the diffeomorphism to the Riemannian manifold in which $\epsilon$ is the step size.

\vspace{-0.1in}
\paragraph{Incorporating structural information.}
We inject the structural information when the received spikes transform into the incoming current of the neuron.
A GNN is leveraged to define the current, conducting message-passing over the graph.
Each node’s representation is derived recursively by neighborhood aggregation \cite{kipf2016semi, velickovic2018graph, hamilton2017inductive}.
Accordingly, $\operatorname{GCN}$ in the proposed neuron is given as follows, 
\begin{align}
    \operatorname{GCN}(\mathbf{s}_i^{l-1}[t];\mathbf{W}^l)&= 
    \operatorname{combine}(\mathbf{s}_i^{l-1}[t], \operatorname{aggregate}(\{\mathbf{s}_j[t]: j \in \Omega_i\};\mathbf{W}^l)),
\end{align}
where  the neighborhood $\Omega_i$ is the set of immediate neighbors centering at node $v_i$.
The $\operatorname{aggregate}$ function aggregates the messages from neighborhood $\Omega_i$, where we create the message of a node by $\mathbf{W}^l\mathbf{s}_j[t]$. 
$\operatorname{combine}(\cdot)$ denotes the combination of  the center node's message and aggregated message.
We utilize GCN \cite{kipf2016semi} as the backbone to define $\operatorname{aggregate}$ and $\operatorname{combine}$.

\vspace{-0.1in}
\paragraph{Diffeomorphism between manifolds.}
In the proposed neuron, we bridge the spikes and manifold representation with the notion of \emph{diffeomorphism} in differential geometry.
A diffeomorphism connects two smooth manifolds, saying $\mathcal{M}$ and $\mathcal{N}$. Formally, a map $f:\mathcal{M}\rightarrow\mathcal{N}$ is a diffeomorphism between $\mathcal{M}$ and $\mathcal{N}$ if the smooth $f$ is bijective and its inverse $f^{-1}$ is also smooth. 

Recall that the tangent space is locally Euclidean. We propose to place the  Euclidean $\mathbf{v}$, a representation of the spikes, in the tangent space $T_{\mathbf{z}}\mathcal M$ of the point $\mathbf{z}$.
In \texttt{MSG}, we choose the \emph{exponential map} to act as the diffeomorphism between the tangent space and  manifold.
With a step size $\epsilon$, we have
\begin{equation}
f (\mathbf{z}_i^{(l-1)}, \epsilon\mathbf{v}_i^{(l-1)}) = \operatorname{Exp}_{\mathbf{z}_i^{(l-1)}}(\epsilon\mathbf{v}_i^{(l-1)}) \in \mathcal M
\end{equation}
Concretely, given $\mathbf{z} \in \mathcal{M}$ and $\mathbf{v} \in T_\mathbf{z}\mathcal{M}$, the exponential map\footnote{The inverse map from $\mathcal M$ to $T_\mathbf{z}\mathcal{M}$ is the logarithmic map.}
 of $\mathbf{v}$ at point $\mathbf{z}$, 
$\operatorname{Exp}_{\mathbf{z}}(\mathbf{v}): T_\mathbf{z}\mathcal{M} \rightarrow \mathcal{M}$,  
maps tangent vector $\mathbf{v}$ onto the manifold  $\mathcal{M}$. 
The map pushes $\mathbf{z}$ along the \emph{geodesic} $\gamma_{\mathbf{z}, \mathbf{v}}(t): [0, 1] \rightarrow \mathcal{M}$ starting at $\gamma_{\mathbf{z}, \mathbf{v}}(0)=\mathbf{z}$ and ending at $\mathbf{y} = \gamma_{\mathbf{z}, \mathbf{v}}(1)$.
$\dot{\gamma}_{\mathbf{z}, \mathbf{v}}(t)$ denotes the velocity of  $\gamma_{\mathbf{z}, \mathbf{v}}(t)$,
and the direction of geodesic at the beginning is given as $\dot{\gamma}_{\mathbf{z}, \mathbf{v}}(0) = \mathbf{v}$.
That is, the tangent vector $\mathbf{v}$, derived from the spikes, pushes the manifold representation along the geodesic via the exponential map.
The advantage of our choice is that we are able to define the diffeomorphism in arbitrary geodesically complete manifold (detailed in Appendix \ref{append. riemannian}).


Note that, our idea is inherently different from the exponential/logarithmic based Riemannian GNNs \cite{chami2019hyperbolic,xiong2022pseudo,liu2019hyperbolica}, which leverage the tangent space of the origin for neighborhood aggregation.
In contrast, we consider the successive process over  the tangent spaces of manifold representations, which will be further studied in Sec. 5.




\vspace{-0.05in}
\paragraph{Model Initialization}

In \texttt{MSG}, we need to simultaneously initialize the spiking input $\mathbf{S}^0 $ and manifold representation $\mathbf{Z}^0$, 
which is a collection of points on the given manifold.
Given a graph $\mathcal{G}(\mathcal V, \mathcal E, \mathbf F, \mathbf A)$, the node features are first encoded by one graph convolution layer $\mathbf{H} = \operatorname{GCN}(\mathbf A, \mathbf F; \mathbf{W}^0)$, and we generate $T$ copies of the node encodings $\mathbf{H}$, where $T$ is the number of time steps in spike trains.
Then, we complete model initialization with the proposed manifold neuron $[\mathbf{S}^0, \mathbf{Z}^0]= \operatorname{MSNeuron}(\mathbf{H}, \mathbf{O})$, 
where the encoding $\mathbf{H}$ is regarded as the incoming current that charges the neuron in each time step.
$\mathbf{O}$ consists of the original points of the manifold, 
e.g., 
in the sphere model of hyperspherical space, the original point is given as the south pole  $\mathbf{o}=[-1,0,...,0]^\top$ and $\mathbf{O}=[\mathbf{o}^\top, ...,\mathbf{o}^\top]^\top$.
Note that, the exponential map in the proposed neuron guarantees that $\mathbf{Z}^0$ lives in the manifold.


\vspace{-0.07in}
\subsection{Learning Approach: Differentiation via Manifold}
\label{sec:backward}
\vspace{-0.03in}

\begin{wrapfigure}{R}{0.52\textwidth}
\begin{minipage}{0.5\textwidth}
\vspace{-0.3in}
\begin{algorithm}[H]
    \caption{Training \texttt{MSG} by the proposed Differentiation via Manifold}
    \label{alg. msg}
    \renewcommand{\algorithmicrequire}{\textbf{Input:}}
     \renewcommand{\algorithmicensure}{\textbf{Output:}}
    \begin{algorithmic}[1]
        \REQUIRE Graph $\mathcal{G}(\mathcal V, \mathcal E, \mathbf F, \mathbf A)$,  Manifold $\mathcal M$,  Loss function over the manifold $\mathcal{L}(\cdot)$, Number of spiking layers $L$, Original points $\mathbf{O}$.
        \ENSURE Parameters $\{\mathbf{W}^l\}_{l=0, \cdots, L}$
        \WHILE{not converging}
        \STATE   \hfill $\rhd$ \emph{forward pass}\\
            \STATE Input current $\mathbf{X}^0 = \operatorname{GCN}(\mathbf A, \mathbf F ; \mathbf{W}^0)$;
            \STATE Initialize $[\mathbf{S}^0, \mathbf{Z}^0] = \operatorname{MSNeuron}(\mathbf{X}^0, \mathbf{O})$;
            \FOR{each  spiking layer $l=1$ to $L$}
            \STATE $\mathbf{X}^{(l-1)} = \operatorname{GCN}(\mathbf A, \mathbf S^{(l-1)} ; \mathbf{W}^l)$;
            \STATE $[\mathbf{S}^l, \mathbf{Z}^l] = \operatorname{MSNeuron}(\mathbf{X}^{(l-1)}, \mathbf{Z}^{(l-1)})$;
            \ENDFOR
        \STATE   \hfill $\rhd$ \emph{backward pass}\\
        \STATE Compute $\nabla_{\mathbf{z}^L} \mathcal{L}$ from $\mathcal{L}(\mathbf{Z}^L)$.
        \FOR{layer $l=L-1$ to 1}
        \STATE Compute $D_{\mathbf{z}^{l}}\psi^{l}, D_{\mathbf{v}^{l-1}}\phi^{l-1}, \frac{\partial \mathbf{v}^{l-1}}{\partial \mathbf{W}^l}$.
        \STATE Compute $\nabla_{\mathbf{z}^l} \mathcal{L}$, $\nabla_{\mathbf{W}^l} \mathcal{L}$ as Eq. \ref{eq. grad}.
        \STATE Update $\mathbf{W}^l$.
        \ENDFOR
        \ENDWHILE
    \end{algorithmic}
\end{algorithm}
\vspace{-0.4in}
\end{minipage}
\end{wrapfigure}


Optimizing SNNs is challenging, as the Heaviside step function is non-differentiable.
In the literature, existing spiking GNNs typically regard SNN as the recurrent neural network, and leverage backward-passing-through-time (BPTT) to train the model \cite{huh2018gradient,neftci2019surrogate, li2021differentiable}. 
Concretely, given a real loss function $\mathcal{L}$, the gradient backpropagation conducts \textbf{Differentiation via Spikes} (\emph{DvS})  $\mathbf s$ as follows,
\begin{align}
    \nabla_{\mathbf{W}^l} \mathcal{L} &= \sum_t [\frac{\partial \mathbf{s}^{l}[t]}{\partial \mathbf{W}^l}]^*\nabla_{\mathbf{s}^l[t]} \mathcal{L},
    \label{DvS}
\end{align}
where $\mathbf W$ is the parameter, and $t$ denotes the time step. 
The surrogate gradient \cite{neftci2019surrogate} is required for $D_{\mathbf{W}^l}\mathbf{s}^l[t]$, where Heaviside step function is replaced by a differentiable surrogate, e.g., sigmoid function.
The differentiation via spikes presents high latency in the backward pass \cite{wu2021training,meng2022training,wu2023tandem}, as it needs to recur all the time steps in BPTT.
We notice that, in the computer vision domain, the sub-gradient method \cite{meng2022training} is proposed to address such issues in Euclidean space. 
However, it cannot be generalized to the Riemannian manifold since the linearity does not hold in Riemannian geometry.


In \texttt{MSG}, we decouple the forward pass and backward pass, and propose \textbf{Differentiation via Manifold} (\emph{DvM})  to avoid the high latency  in differentiation via spikes.
The overall procedure of training \texttt{MSG} by the proposed learning approach is summarized in Algorithm \ref{alg. msg}.
Thanks to the parallel forwarding of spikes and manifold representation, 
the proposed neuron provides us with an alternative of studying $\nabla_{\mathbf{W}^l} \mathcal{L}$ through the forwarding pass on the manifold (i.e., differentiation via manifold).
Nevertheless, \emph{it is nontrivial and it requires to derive the pullback between different dual spaces.}

\vspace{-0.05in}
\paragraph{Pushforward, Pullback and Dual Space.}
We first introduce the differentiation in Riemannian geometry which is essentially different from that in Euclidean space.
In Riemannian geometry, a \emph{pushforward} refers to a derivative of a map connecting two manifolds $\mathcal M$ and $\mathcal N$.
 Concretely, given $f: \mathcal M \to \mathcal N$ and a point $\mathbf{z} \in \mathcal{M}$, 
 the pushforward $D_\mathbf{z}f$ maps a tangent vector $\mathbf{v} \in T_\mathbf{z}\mathcal M$ to the tangent vector $D_\mathbf{z}f(\mathbf{v}) \in T_{f(\mathbf{z})}\mathcal{N}$. 
 On the notation, for a manifold-valued function $f(\mathbf{z})=\mathbf{p} \in \mathcal{N}$, $\partial \mathbf{p}/\partial \mathbf{z}$ is equivalent to $D_\mathbf{z}f$. 
 For a scalar function $f$, $D_\mathbf{z}f$ is interchangeable with $\nabla_\mathbf{z}f$.

In the proposed \texttt{MSG}, we consider a scalar loss function on the manifold $\mathcal{L}: \mathcal{M}\rightarrow \mathbb{R}$.
The pushforward $D_\mathbf{z}\mathcal{L}$ at point $\mathbf{z}\in \mathcal{M}$ maps tangent vector $\mathbf{v} \in T_\mathbf{z}\mathcal M$ to a scalar value and,
correspondingly, 
$D_\mathbf{z}\mathcal{L}$ belongs to the \emph{dual space} of the tangent space $T_\mathbf{z}^*\mathcal{M}$, 
which is a vector space consisting all linear functional $F:T_\mathbf{z}\mathcal{M}\rightarrow \mathbb{R}$.
As the tangent spaces at different points of the manifold are different,
it requires a \emph{pullback} that maps the dual space $T_{\mathbf{z}^l}^*\mathcal{M}$  to the dual space $T_{\mathbf{z}^{(l+1)}}^*\mathcal{M}$.

We derive the backward gradient with properties of differential $1-$form (Lemma \ref{lemma. form}), communication (Lemma \ref{lemma. commu}), and pullback of a sum and a product (Lemma \ref{lemma. sum}) detailed in Appendix \ref{proof. grad}.


\begin{theorem}[Backward Gradient]
\label{grad}
Let $\mathcal{L}$ be the scalar-valued function, and $\mathbf{z}^l$ is the output of $l$-th layer with parameter $\mathbf{W}^l$, which is delivered by tangent vector $\mathbf{v}^l$. Then, the gradient of  function $\mathcal{L}$ w.r.t $\mathbf{W}^l$ is given as follows:
     \vspace{-0.03in}
\begin{align}
\label{eq. grad}   
    \nabla_{\mathbf{W}^l} \mathcal{L} = [\frac{\partial \mathbf{v}^{l-1}}{\partial \mathbf{W}^l}]^* [D_{\mathbf{v}^{l-1}}\phi^{l-1}]^* \nabla_{\mathbf{z}^l} \mathcal{L},    \quad
     \nabla_{\mathbf{z}^l} \mathcal{L} = [D_{\mathbf{z}^{l}}\psi^{l}]^* \nabla_{\mathbf{z}^{l+1}} \mathcal{L}, 
     \vspace{-0.03in}
\end{align}
where $\phi^{l-1}(\cdot)=\operatorname{Exp}_{\mathbf{z}^{l-1}}(\cdot)$, $\psi^{l}(\cdot)=\operatorname{Exp}_{(\cdot)}({\mathbf{v}^{l}})$, and $[\cdot]^*$ means the matrix form of pullback.
\end{theorem}
The detailed proof is given in Appendix \ref{proof. grad}, and we derive the two Jacobian matrices $D_{\mathbf{v}^{l-1}}\phi^{l-1}$ and $D_{\mathbf{z}^{l}}\psi^{l}$ in Appendix \ref{formulas}.
There are three key advantages of the proposed \emph{DvM}.
First,  every term in Equation (\ref{eq. grad}) is \textbf{differentiable}, and thereby the surrogate gradient is no longer needed.
Second,  \emph{DvM} enables the \textbf{recurrence-free} backward pass alleviating the high latency training.  
We specify that both \emph{DvM} and the previous \emph{DvS} recurrently compute every time step in the forward pass, and the difference lies in the backward pass.
In particular, we only conduct recurrence-free gradient backpropagation layer by layer, while the previous \emph{DvS} recurs every time step of each layer in BPTT.
In addition to the differentiable and recurrence-free properties, \emph{DvM} does not suffer from gradient vanishing/explosion, and the empirical evidence is provided in Appendix \ref{append.add}.



\vspace{-0.1in}
\section{Theory: MSG as Neural ODE Solver}
\vspace{-0.05in}

Next, we demonstrate the theoretical aspects of our model that \emph{\texttt{MSG} approximates a solver of manifold Ordinary Differential Equations (ODEs).}

\begin{wrapfigure}{l}{0.505\textwidth}
\centering
 \vspace{-0.05in}
\includegraphics[width=1\linewidth]{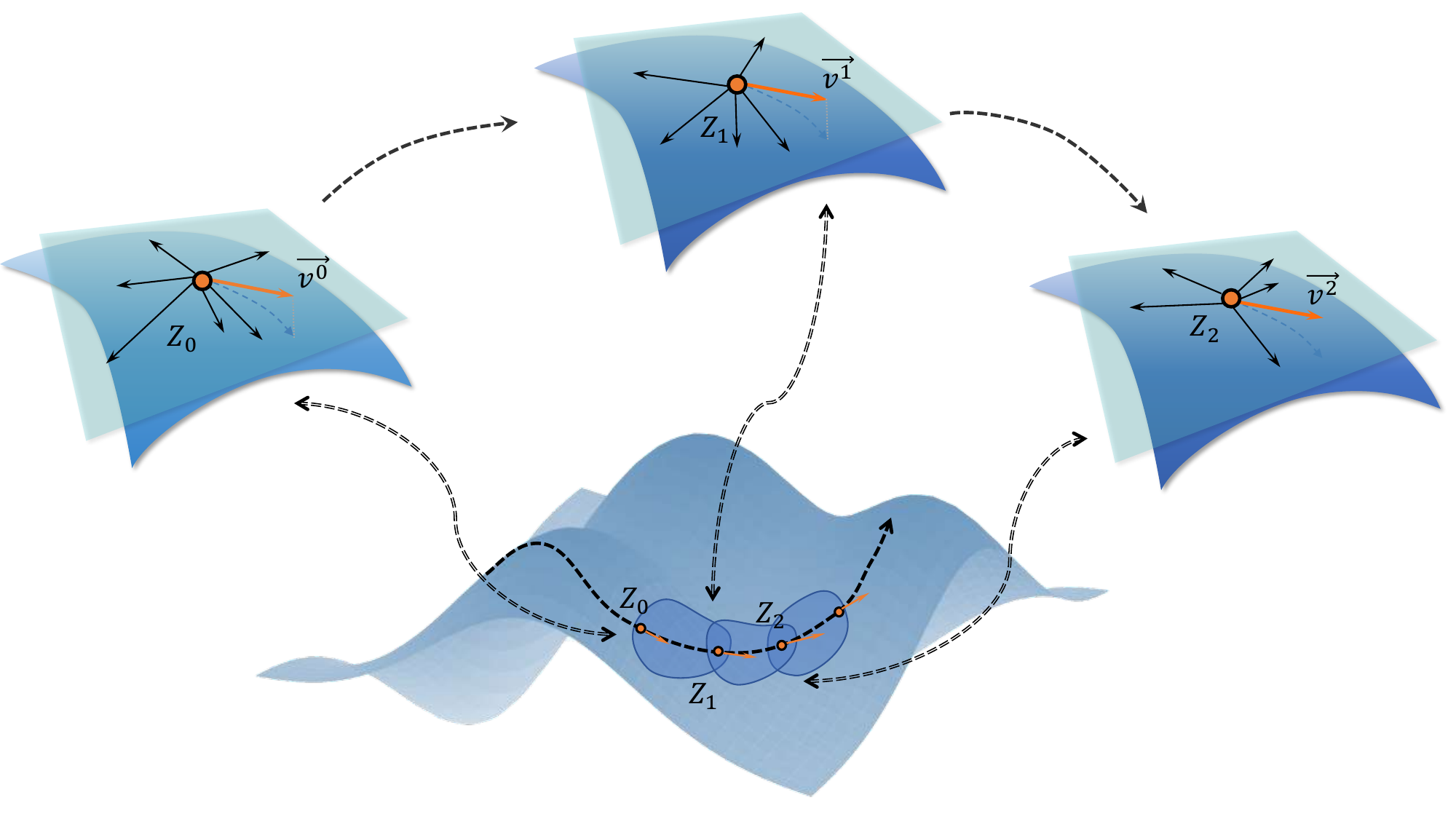}
 \vspace{-0.15in}
\caption{Charts given by the logarithmic map.}
\vspace{-0.05in}
\label{fig. dynamicChart}
\end{wrapfigure}

We leverage the notion of \textbf{chart} to study the relationship between \texttt{MSG} and manifold ODE.
A  \emph{manifold ODE} defined as 
\begin{equation}
\frac{d\mathbf{z}(t)}{dt} = u(\mathbf{z}(t), t), \quad  \mathbf{z}(t) \in \mathcal M,
\label{mode}
\end{equation}
describes a \emph{vector field} $u$ that maps a smooth path $ \mathbf z(t): [0,1] \to \mathcal M$ to the tangent bundle $T\mathcal M$ 
\footnote{The tangent bundle $T\mathcal M$  is the disjoint union of all the tangent spaces of the manifold $T\mathcal M=\bigsqcup_{\mathbf z \in \mathcal M}T_{\mathbf z}\mathcal M$.}.
In other words, the vector field $u$ assigns each point $\mathbf{z}(t) \in \mathcal M$ to a tangent vector $u(\mathbf{z}(t), t) \in T_{\mathbf{z}(t)}\mathcal{M}$.
A \emph{chart} at $\mathbf z$, denoted as $(U_\mathbf{z}, \phi_\mathbf{z})$,  is a smooth bijection $\phi_\mathbf{z}$ between $\mathbf z$'s neighborhood $U_\mathbf z \subset \mathcal M$ and a subspace of Euclidean space.
Thus, the chart relates Eq. (\ref{mode})  to ODE in Euclidean space \cite{lou2020neural}. 
Specifically, 
    if $\mathbf{y}(t): [\tau, \tau + \epsilon] \to \mathbb R^n$ is the solution of 
         \vspace{-0.03in}
    \begin{align}
        \frac{d\mathbf{y}(t)}{dt} = (D_{\phi_i^{-1}(\mathbf{y}(t))}\phi_i)u(\phi_i^{-1}(\mathbf{y}(t)), t),
        \vspace{-0.03in}
    \end{align}
    then $\mathbf{y}(t)=\phi_i(\mathbf{z}(t))$ is a valid solution of Eq. (\ref{mode}) on $t \in [\tau, \tau + \epsilon]$.
\begin{definition}[Dynamic Chart Solver \cite{lou2020neural}]
    \label{def. dynamic}
 The manifold ODE in Eq. (\ref{mode}) with initial condition $\mathbf{z}(0)=\mathbf{z}$
 can be solved with a finite collection of successive charts $\{(U_i, \phi_i)\}_{i=1,...,L}$. 
If $\operatorname{ode}_i$ is the numerical solver to Euclidean ODE corresponding to the $i$-th chart, $\mathbf y(t)=\operatorname{ode}_i(t)$ on $ [\tau_i, \tau_i + \epsilon_i]$,
then $\mathbf{z}(t)$ in Eq. (\ref{mode}) is given as  
    \begin{align}
(\phi_L^{-1} \circ \operatorname{ode}_L \circ (\phi_L \circ \phi_{L-1}^{-1}) \circ ... \circ (\phi_2\circ \phi_1^{-1}) \circ \operatorname{ode}_1 \circ \phi_1)(t).
\label{dynamicChartSolver}
\vspace{-0.03in}
    \end{align}
\end{definition}
That is, a manifold ODE can be solved in Euclidean subspaces given by a series of successive charts.


In \texttt{MSG}, we consider the charts given by the logarithmic map as illustrated in \ref{fig. dynamicChart},
and we prove that \texttt{MSG} approximates a dynamic chart solver of manifold ODE (Theorem \ref{ode}).

\begin{theorem}[\texttt{MSG} as Dynamic Chart Solver]
\label{ode}
    If $\mathbf{y}(t): [\tau, \tau + \epsilon] \to \mathbb R^n$ is the solution of 
         \vspace{-0.03in}
   \begin{align}
    \label{eq. Eode}
        \frac{d\mathbf{y}(t)}{dt} = (D_{\operatorname{Exp}_\mathbf{z}(\mathbf{y(t)})} \operatorname{Log}_\mathbf{z}) u(\operatorname{Exp}_\mathbf{z}(\mathbf{y}(t)), t),
            \vspace{-0.05in}
    \end{align}
    then 
  $\mathbf{z}(t)=\operatorname{Exp}_\mathbf{z}(\mathbf{y}(t))$ 
  is a valid solution to the manifold ODE of Eq. (\ref{mode}) on $t \in [\tau, \tau + \epsilon]$, where $\mathbf{z}=\mathbf{z}(\tau)$.
 If $\mathbf{y}(t)$ is given by the first-order approximation with the $\epsilon$ small enough,
  \vspace{-0.03in}
    \begin{align}
        \mathbf{y}(\tau + \epsilon)=\epsilon \cdot(D_\mathbf{z}\operatorname{Log}_\mathbf{z})u(\mathbf{z}(\tau), \tau),
            \vspace{-0.05in}
    \end{align}
    then the update process of  Eqs. (\ref{eq. x}) and  (\ref{eq. z}) in \texttt{MSG} is equivalent to Dynamic Chart Solver in Eq. (\ref{dynamicChartSolver}).
     \vspace{-0.03in}
\end{theorem}
    \vspace{-0.05in}
\begin{proof}
    The proof utilizes some facts of Riemannian manifolds and is detailed in Appendix \ref{proof. ode}.
\end{proof}
    \vspace{-0.05in}
In other words, in \texttt{MSG}, the transformation of manifold input and output is described as some manifold ODE,
whose vector field is governed by a spiking-related neural network in the tangent bundle.
To solve the manifold ODE, \texttt{MSG} leverages the Dynamic Chart Solver (Definition \ref{def. dynamic}).
Specifically, each manifold spiking layer corresponds to a chart, and thus the number of spiking layers equals to the number of charts.
Each layer solves the ODE of a smooth path $\mathbf{y}(t): [\tau, \tau + \epsilon] \to \mathbb R^n$ in the tangent space that centered at the manifold layer input. 
With the first-order approximation in Theorem \ref{ode}, given a step size $\epsilon$, the endpoint $\mathbf{y}(\tau + \epsilon)$ of the path is parameterized by a GNN related to the spikes.
 Layer-by-layer forwarding solves the manifold ODE from the current chart to the successive chart.
 Consequently, the manifold output of \texttt{MSG} approximates the solution to the manifold ODE.

 We notice that a recent work \cite{yin2024continuous} connects spiking GNN to an ODE in Euclidean space.
 In contrast, the proposed \texttt{MSG} is essentially related to the manifold ODE. 


% ...

 The \textbf{Appendix} contains the proofs, the derivation of Jacobian, necessary facts on Riemannian geometry (i.e., Lorentz/Sphere model, stereographic projection and $\kappa$-stereographic model, and Cartesian product and product space), empirical details and additional results.

\vspace{-0.12in}
\section{Experiments}
\label{exp}

            \vspace{-0.05in}
We conduct extensive experiments with $12$ strong baselines to evaluate the proposed \texttt{MSG} in terms of  (1) the representation effectiveness, (2) the energy efficiency, and (3) the advantages of the proposed components. Additional results are presented in Appendix \ref{append.add}.

\vspace{-0.1in}
\subsection{Experimental Setups}
\label{exp_setup}

    \vspace{-0.03in}
\paragraph{Datasets \& Baselines.}
Our experiments are conducted on $4$ commonly used benchmark datasets including two popular co-purchase graphs: \emph{Computers} and \emph{Photo}\cite{shchur2018pitfalls}, and two co-author graphs: \emph{CS} and \emph{Physics} \cite{shchur2018pitfalls}.
We compare the proposed \texttt{MSG} with $12$ strong baselines of three categories:
(1) \emph{ANN-based Euclidean GNNs}: the popular GCN \cite{kipf2016semi}, GAT \cite{velickovic2018graph}, GraphSAGE \cite{hamilton2017inductive} and SGC \cite{wu2019simplifying},
(2) \emph{ANN-based Riemannian GNNs}: HGCN \cite{chami2019hyperbolic} and HyboNet \cite{chen2021fully} of hyperbolic spaces, $\kappa-$GCN \cite{bachmann2020constanta} of the constant curvature space, and the recent $Q-$GCN \cite{xiong2022pseudo} of the quotient space,
(3) \emph{The previous Euclidean Spiking GNNs}: SpikeNet \cite{li2023scaling}, SpikeGCN \cite{zhu2022spiking}, SipkeGraphormer \cite{sun2024spikegraphormer} (termed as SpikeGT for short) and the recent SpikeGCL \cite{li2023graph}.
Note that, we focus on the graph representation learning on static graphs, and thereby graph models for the dynamic ones are out of the scope of this paper.
SpikeNet \cite{li2023scaling} was originally designed for dynamic graphs, and we utilize its variant for static graphs according to \cite{li2023graph}.
So far, spiking GNN has not yet been connected to Riemannian manifolds, and we are devoted to bridging this gap.
Datasets/baselines are detailed in Appendix \ref{append. exp}.

    \vspace{-0.1in}
\paragraph{Evaluation Protocol.}
All models are evaluated by node classification and link prediction tasks.
The evaluation metrics are classification accuracy and Area Under Curve (AUC) for link prediction, respectively.
The hyperparameter setting is the same as the original papers.
We perform $10$ independent runs for each case, and report the mean with standard derivations.
Experiments are conducted on the hardware of NVIDIA GeForce RTX 4090 GPU 24GB memory, and AMD EPYC 9654 CPU with 96-Core Processor.
Our model is built upon GeoOpt \cite{geoopt19}, SpikingJelly \cite{geoopt19} and PyTorch \cite{paszke2019pytorcha}.

    \vspace{-0.1in}
\paragraph{Model Instantiation \& Configuration.}
Note that, the proposed \texttt{MSG} applies to any Constant Curvature Space (CCS) or the product of CCS. 
We instantiate \texttt{MSG} in the Lorentz model of hyperbolic space by default (whose Riemannian metric, exponential map, and \underline{the derived Jacobian} is given in Appendix \ref{formulas}), and study the impact of representation space in the Ablation Study.
The dimension of the representation space is set as $32$.
The manifold spiking neuron is based on the IF model \cite{burkitt2006review} by default, and it is ready to switch to the LIF model \cite{burkitt2006review} whose results are given in Appendix \ref{append.add}. The time steps $T$ for neurons is set to $5$ or $15$. The step size $\epsilon$ in Eq. \ref{eq. z} is set  to $0.1$.
The hyperparameters are tuned with grid search, in which the learning rate is $\{0.01, 0.003\}$ for node classification and $\{0.003, 0.001\}$ for link prediction, and the dropout rate is in $\{0.1, 0.3, 0.5\}$. 
We provide the source code of \texttt{MSG} at the anonymous link \url{https://anonymous.4open.science/r/MSG-16E9}.

\begin{table}[t]
    \caption{Node Classification (NC) in terms of classification accuracy (\%) and Link Prediction in terms of AUC (\%) on Computers, Photo, CS and Physics datasets. The best results are \textbf{boldfaced}, and the runner-ups are \underline{underlined}. The standard derivations are given in the subscripts.}
    \resizebox{1\linewidth}{!}{
    \rowcolors{11}{orange!15}{orange!15} 
    \begin{tabular}{cl|cc|cc|cc|cc}
    \hline 
        &  & \multicolumn{2}{c|}{\textbf{Computers}}  & \multicolumn{2}{c|}{\textbf{Photo}}  & \multicolumn{2}{c|}{\textbf{CS}} & \multicolumn{2}{c}{\textbf{Physics}} \\
       & & NC & LP & NC & LP  & NC & LP & NC & LP  \\ 
    \hline
     \multirow{4}{*}{\rotatebox{90}{\footnotesize ANN-\textbf{E}}} 
       & GCN \cite{kipf2016semi} 
       & 83.55{\scriptsize$\pm$0.71} & 92.07{\scriptsize$\pm$0.40} & 86.01{\scriptsize$\pm$0.20} & 88.84{\scriptsize$\pm$0.39} & {91.68\scriptsize$\pm$0.84} & 93.68{\scriptsize$\pm$0.84} & 95.03{\scriptsize$\pm$0.19} & 93.46{\scriptsize$\pm$0.39} \\ 
       & GAT \cite{velickovic2018graph} 
       & 86.82{\scriptsize$\pm$0.04} & 91.91{\scriptsize$\pm$1.08} & 86.68{\scriptsize$\pm$1.32} & 88.45{\scriptsize$\pm$0.07} & 91.74{\scriptsize$\pm$0.22} & 94.06{\scriptsize$\pm$0.70} & 95.11{\scriptsize$\pm$0.29} & 93.44{\scriptsize$\pm$0.70}  \\ 
       & SGC \cite{wu2019simplifying} 
       & 82.17{\scriptsize$\pm$1.25} & 90.46{\scriptsize$\pm$0.80} & 87.91{\scriptsize$\pm$0.65} & 89.84{\scriptsize$\pm$0.40} & 92.09{\scriptsize$\pm$0.05} & \textbf{95.94{\scriptsize$\pm$0.43}} & 94.77{\scriptsize$\pm$0.32} & \underline{95.93{\scriptsize$\pm$0.70}}  \\ 
       & SAGE \cite{hamilton2017inductive} 
       & 81.69{\scriptsize$\pm$0.86} & 90.56{\scriptsize$\pm$0.48} & 89.41{\scriptsize$\pm$1.28} & 89.86{\scriptsize$\pm$0.90} & \textbf{92.71{\scriptsize$\pm$0.73}} & 95.22{\scriptsize$\pm$0.14} & 95.62{\scriptsize$\pm$0.26} & 95.75{\scriptsize$\pm$0.37}  \\ 
    \hline
       \multirow{4}{*}{\rotatebox{90}{\footnotesize ANN-\textbf{R}}} 
       & HGCN \cite{chami2019hyperbolic} 
       & 88.71{\scriptsize$\pm$0.24} & \underline{96.88{\scriptsize$\pm$0.53}} & 89.18{\scriptsize$\pm$0.50} & 94.54{\scriptsize$\pm$0.20} & 90.72{\scriptsize$\pm$0.16} & 93.02{\scriptsize$\pm$0.26} & 94.46{\scriptsize$\pm$0.20} & 94.10{\scriptsize$\pm$0.64}  \\ 
       & $\kappa$-GCN \cite{bachmann2020constanta} 
       & \underline{89.20{\scriptsize$\pm$0.50}} & 95.30{\scriptsize$\pm$0.20} & 92.22{\scriptsize$\pm$0.62} & 94.89{\scriptsize$\pm$0.15} & 91.98{\scriptsize$\pm$0.15} & 94.86{\scriptsize$\pm$0.18} & \underline{95.85{\scriptsize$\pm$0.20}} & 94.58{\scriptsize$\pm$0.22}  \\ 
       & $\mathcal{Q}$-GCN \cite{xiong2022pseudo} 
       & 85.94{\scriptsize$\pm$0.93} & \textbf{96.98{\scriptsize$\pm$0.05}} & 92.50{\scriptsize$\pm$0.95} & \underline{97.47{\scriptsize$\pm$0.03}} & 91.18{\scriptsize$\pm$0.28} & 93.39{\scriptsize$\pm$0.20} & 94.84{\scriptsize$\pm$0.25} & OOM  \\ 
       & HyboNet \cite{chen2021fully} 
       & 86.29{\scriptsize$\pm$2.30} & 96.80{\scriptsize$\pm$0.05} & 92.67{\scriptsize$\pm$0.09} & \textbf{97.70{\scriptsize$\pm$0.07}} & 92.34{\scriptsize$\pm$0.03} & \underline{95.65{\scriptsize$\pm$0.26}} & 95.56{\scriptsize$\pm$0.18} & \textbf{98.46{\scriptsize$\pm$0.49}}  \\ 
    \hline    
       & SpikeNet \cite{li2023scaling} 
       & 88.00{\scriptsize$\pm$0.70} & - & \underline{92.90{\scriptsize$\pm$0.10}} & - & 92.15{\scriptsize$\pm$0.18} & - & 92.66{\scriptsize$\pm$0.30} & -   \\ 
       & SpikeGCN \cite{zhu2022spiking} 
       & 86.90{\scriptsize$\pm$0.30} & 91.12{\scriptsize$\pm$1.79} & 92.60{\scriptsize$\pm$0.70} & 93.84{\scriptsize$\pm$0.03} & 90.86{\scriptsize$\pm$0.11} & 95.07{\scriptsize$\pm$1.22} & 94.53{\scriptsize$\pm$0.18} & 92.88{\scriptsize$\pm$0.80}   \\ 
       & SpikeGCL \cite{li2023graph} 
       & 89.04{\scriptsize$\pm$0.08} & 92.72{\scriptsize$\pm$0.03} & 92.50{\scriptsize$\pm$0.17} & 95.58{\scriptsize$\pm$0.11} & 91.77{\scriptsize$\pm$0.11} & 95.13{\scriptsize$\pm$0.24} & 95.21{\scriptsize$\pm$0.10} & 94.15{\scriptsize$\pm$0.29}   \\ 
       \multirow{-4}{*}{\rotatebox{90}{\footnotesize SNN-\textbf{E}}} 
       & SpikeGT \cite{sun2024spikegraphormer} 
       & 81.00{\scriptsize$\pm$1.06} & - & 90.66{\scriptsize$\pm$0.38} & - & 91.86{\scriptsize$\pm$0.61} & - & 94.38{\scriptsize$\pm$1.57} & -  \\
    \hline
       & \texttt{MSG} (Ours) 
       & \textbf{89.27{\scriptsize$\pm$0.19}} & 94.65{\scriptsize$\pm$0.73} & \textbf{93.11{\scriptsize$\pm$0.11}}  & 96.75{\scriptsize$\pm$0.18} 
       & \underline{92.65{\scriptsize$\pm$0.04}} & 95.19{\scriptsize$\pm$0.15} & \textbf{95.93{\scriptsize$\pm$0.07}} & 93.43{\scriptsize$\pm$0.16}  \\
       \hline
    \end{tabular}
    }
    \label{table:NC_LP}
    \vspace{-0.2in}
\end{table}

    \vspace{-0.07in}
\subsection{Results \& Discussion}

    \vspace{-0.07in}
\paragraph{Effectiveness.}
We evaluate the effectiveness of \texttt{MSG} in both node classification and link prediction tasks.
Specifically, for node classification, we cannot directly feed the manifold representations of Riemannian baselines to a softmax layer with Euclidean measure.
We suggest to bridge the manifold representation and Euclidean softmax with the logarithmic map of respective manifold, described in Appendix \ref{append. riemannian}.
For link prediction, we utilize the generalized sigmoid for all the baselines, i.e., the Fermi-Dirac decoder \cite{chami2019hyperbolic} in which the distance function is defined under the respective geometry.
The performance of both learning tasks on Computer, Photo, CS and Physics datasets are collected in Table \ref{table:NC_LP}.
Note that, SpikeNet and SpikeGT cannot do link prediction, since they are designed for node classification and do not offer spiking representation.
\emph{The proposed \texttt{MSG} consistently  achieves the best results among SNN-based models.}
In addition, \texttt{MSG} generally outperforms the best ANN-based baselines in node classification, and has competitive results to the recent ANN-based Riemannian baselines in link prediction.

\begin{wraptable}{r}{0.6\textwidth}
 \vspace{-0.25in}
\centering
\caption{Ablation study of geometric variants. Results of node classification in terms of ACC (\%).}
\vspace{0.03in}
\resizebox{1\linewidth}{!}{
\begin{tabular}{c| c c c c}
  \hline
 & Computers  & Photo      & CS         & Physics    \\
  \hline
$\mathbb{H}^{32}$ & \textbf{89.27{\scriptsize$\pm$0.19}} & \textbf{93.11{\scriptsize$\pm$0.11}} & 92.65{\scriptsize$\pm$0.04} & \textbf{95.93{\scriptsize$\pm$0.07}}\\ 
$\mathbb{S}^{32}$ & 87.84{\scriptsize$\pm$0.77} & 92.03{\scriptsize$\pm$0.79} & \underline{92.72{\scriptsize$\pm$0.06}} & 95.85{\scriptsize$\pm$0.02}\\ 
$\mathbb{E}^{32}$ & 88.94{\scriptsize$\pm$0.24} & \underline{92.93{\scriptsize$\pm$0.21}} & \textbf{92.82{\scriptsize$\pm$0.04}} & 95.81{\scriptsize$\pm$0.04}\\ 
\hline
$\mathbb{H}^{16}\times \mathbb{H}^{16}$& \underline{89.18{\scriptsize$\pm$0.25}} & 92.06{\scriptsize$\pm$0.14} & 92.67{\scriptsize$\pm$0.10} &\underline{95.90{\scriptsize$\pm$0.04}} \\ 
$\mathbb{H}^{16}\times \mathbb{S}^{16}$ & 88.00{\scriptsize$\pm$1.05} & 91.97{\scriptsize$\pm$0.08} & 92.33{\scriptsize$\pm$0.21} & 95.73{\scriptsize$\pm$0.11}\\ 
$\mathbb{S}^{16} \times \mathbb{S}^{16}$ & 82.49{\scriptsize$\pm$1.18} & 92.31{\scriptsize$\pm$0.45} & 92.18{\scriptsize$\pm$0.21} & 95.81{\scriptsize$\pm$0.10 }\\
\hline
\end{tabular}
\vspace{-0.2in}
}
\end{wraptable}

\vspace{-0.07in}
\paragraph{Ablation Study.}
Here, we examine the impact of representation space and the effectiveness of the proposed \emph{Differentiation via Manifold (DvM)}.
For the former goal, we instantiate $6$ geometric variants of \texttt{MSG} in hyperbolic space $\mathbb H^{32}$, hyperspherical space $\mathbb S^{32}$, Euclidean space $\mathbb E^{32}$ and the products of $\mathbb H^{16} \times \mathbb H^{16}$, $\mathbb H^{16} \times \mathbb S^{16}$ and $\mathbb S^{16} \times \mathbb S^{16}$.
The superscript denotes the dimension of representation space, and we leverage \emph{DvM} for optimization.
Manifold variants generally achieve superior results to the Euclidean one, thus verifying our motivation.
On CS dataset, the performance of geometric variants is aligned with that of Euclidean and Riemannian baselines in Table \ref{table:NC_LP}. 
The proposed \texttt{MSG}  is ready to switch among $\mathbb H$, $\mathbb S$,  $\mathbb E$, and their products, matching the geometry of graphs.


\begin{figure}[h]
\vspace{-0.07in}
    \centering
    \subfigure[Backward times in model training.]{
        \includegraphics[width=0.48\textwidth]{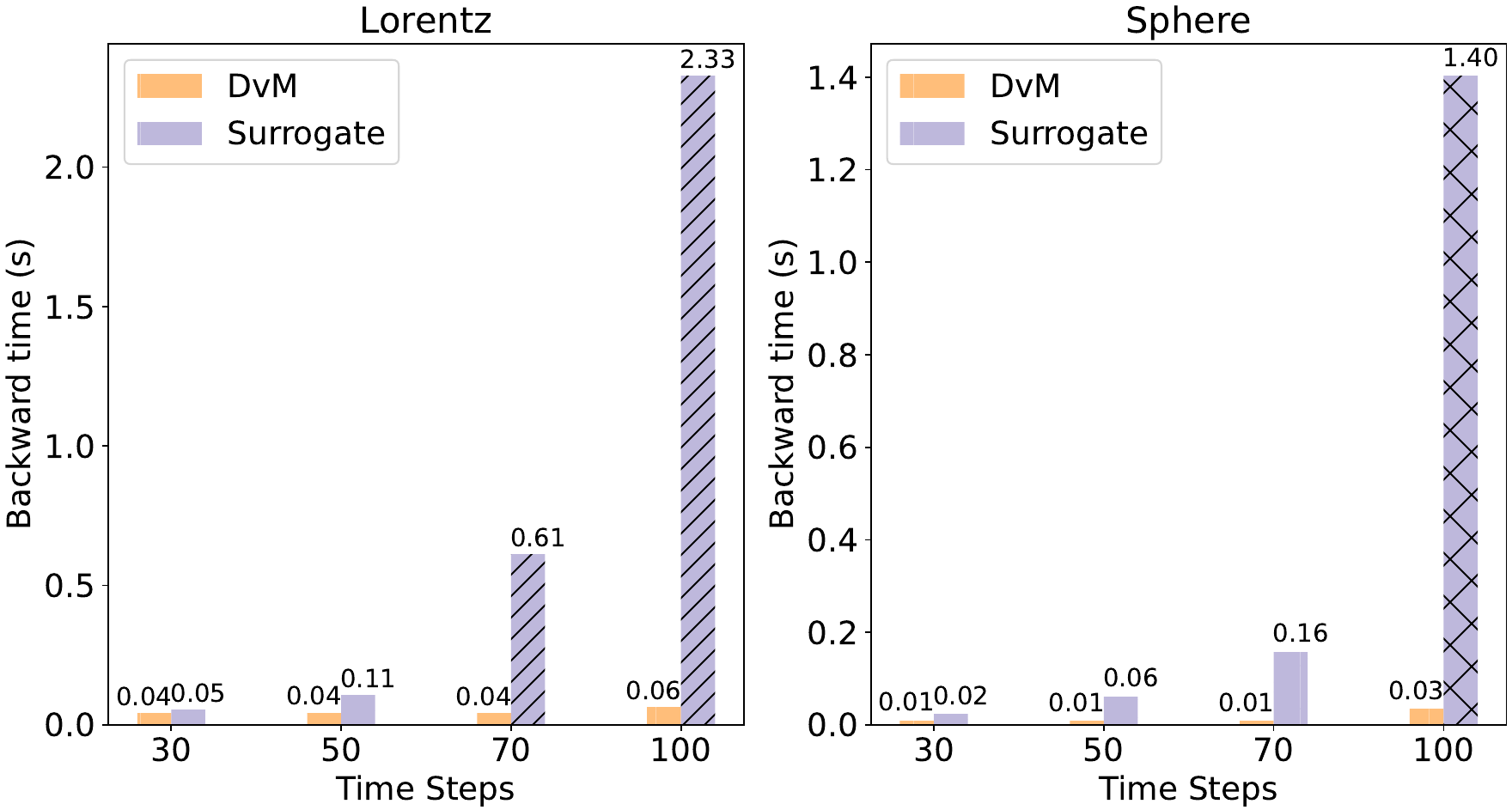}
                 \label{fig:fig_micPerMon_time}
    }
        \subfigure[Gradient norm of $L$ regarding tangent vector $\mathbf{v}$.]{
        \includegraphics[width=0.475\textwidth]{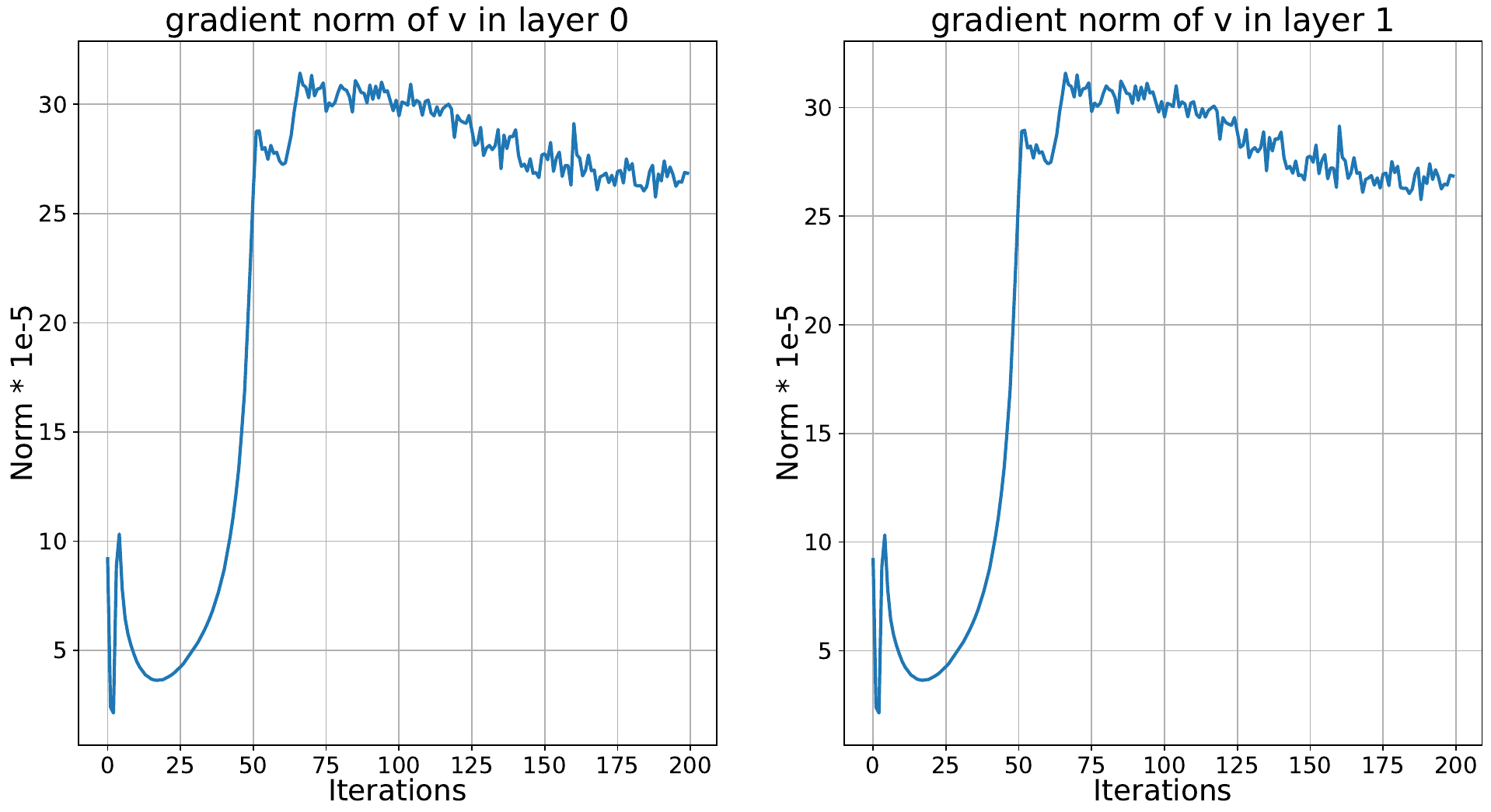}
           \label{fig:fig_micPerMon_norm}
    }
    \vspace{-0.07in}
    \caption{Backward time and gradient norm for node classification on Computer.}
    \vspace{-0.07in}
    \label{fig:fig_micPerMon}
\end{figure}

 To examine the effectiveness of \emph{DvM}, we design the optimization variant (named as Surrogate) for a given representation space. 
 In the variant, we conduct differentiation via spikes and leverage BPTT for optimization,  same as previous spiking GNNs.
The training time of the optimization variants in different representation spaces are given in Fig. \ref{fig:fig_micPerMon_time}.
Backward time of \emph{DvM} is significantly less than that of BPTT algorithm.
The reason is that \emph{DvM} no longer needs recurrent gradient calculation of each time step (\underline{recurrence-free}), while BPTT leads to high training time especially when the time step is large.
In addition, we examine the backward gradient of \emph{DvM}, and plot the gradient norm of each layer in Fig. \ref{fig:fig_micPerMon_norm}.
It demonstrates that \emph{DvM} does not suffer from gradient vanishing/explosion.

 \begin{table}[t]
    \centering
    \caption{Energy cost. The number of parameters at the running time (KB) and theoretical energy consumption (mJ) on Computers, Photo, CS and Physics datasets. The best results are \textbf{boldfaced}, and the runner ups are \underline{underlined}. }
    \vspace{-0.05in}
    \centering
    \resizebox{0.9\linewidth}{!}{
    \begin{tabular}{c l  |  r >{\columncolor{orange!15}}r   |r >{\columncolor{orange!15}} r|  r >{\columncolor{orange!15}} r | r >{\columncolor{orange!15}} r }
    \hline
       &  & \multicolumn{2}{c}{\textbf{Computers}}  & \multicolumn{2}{|c}{\textbf{Photo}}  & \multicolumn{2}{|c}{\textbf{CS}} & \multicolumn{2}{|c}{\textbf{Physics}} \\
       & & \#(para.) & energy & \#(para.) & energy   & \#(para.) & energy & \#(para.) & energy \\ 
     \hline
        \multirow{4}{*}{\rotatebox{90}{\footnotesize ANN-\textbf{E}}} 
        & GCN \cite{kipf2016semi}
        & \underline{24.91}     & 1.671      & \underline{24.14}     & 0.893      & 218.29   & 18.444     & 269.48   & 42.842 \\ 
        & GAT \cite{velickovic2018graph}
        & 24.99     & 2.477      & 24.22     & 1.273      & 218.38   & 28.782     & 269.55   & 81.466  \\ 
        & SGC \cite{wu2019simplifying}
        & \textbf{7.68}      & 0.508      & \textbf{5.97}      & 0.219      & \textbf{102.09}   & 8.621      & \textbf{42.08}    & 6.688 \\ 
        & SAGE \cite{hamilton2017inductive}
        & 49.77     & 1.671      & 48.23     & 0.893      & 436.53   & 18.444     & 538.92   & 42.842  \\ 
    \hline
       \multirow{4}{*}{\rotatebox{90}{\footnotesize ANN-\textbf{R}}} 
        & HGCN \cite{chami2019hyperbolic}
        & 24.94     & 1.614      & 24.96     & 0.869      & \underline{217.79}    & 18.390      & 269.31    & 42.800 \\ 
        &$\kappa$-GCN \cite{bachmann2020constanta}
        & 25.89     & 1.647      & 25.12     & 0.889      & 218.24   & 18.440      & 269.44   & 42.836 \\ 
        & $\mathcal{Q}-$GCN \cite{xiong2022pseudo}
        & 24.93     & 1.629      & 24.96     & 0.876      & 217.83   & 18.393     & 269.34   & 42.809 \\ 
        & HyboNet \cite{chen2021fully} 
        & 27.06     & 1.625      & 26.29     & 0.875      & 219.94   & 18.399     & 271.47   & 42.825 \\ 
    \hline
        \multirow{4}{*}{\rotatebox{90}{\footnotesize SNN-\textbf{E}}} 
        & SpikeNet \cite{xu2021exploiting}
        & 101.22    & \underline{0.070}      & 98.07     & \textbf{0.040}     & 438.51    & 0.218      & 540.04    & 0.334  \\ 
        & SpikingGCN \cite{zhu2022spiking}
        & 38.40     & 0.105      & 29.84     & 0.046      & 510.45    & 1.871      & 210.40     & 1.451  \\ 
        & SpikeGCL \cite{li2023graph}
        & 59.26     & 0.121      & 57.85     & 0.067      & 445.69    & \underline{0.128}      & 548.74    & \underline{0.214} \\ 
        & SpikeGT \cite{sun2024spikegraphormer}
        & 77.07     & 1.090      & 74.46     & 0.584      & 365.28    & 6.985      & 355.77    & 12.524  \\ 
    \hline
        & \texttt{MSG}(Ours)  
        & 26.95     & \textbf{0.047}      & 25.68     & \underline{0.043}     & 226.15   & \textbf{0.026}      & \underline{143.72}    & \textbf{0.029}  \\ 
    \hline
    \end{tabular}
    }
        \vspace{-0.22in}
    \label{table. energy}
\end{table}

     \vspace{-0.12in}
\paragraph{Energy Cost.}
We investigate the energy cost of the graph models in terms of theoretical energy consumption (mJ) \cite{zhu2022spiking, li2023graph}, whose formula is specified  in Appendix \ref{append. exp}.
We summarize the results  for node classification in Table \ref{table. energy} in which the number of parameters at the running time is listed as a reference.
It shows that SNN-based models generally enjoy less energy cost than ANN-based ones.
Note, \texttt{MSG} achieves the best energy efficiency among SNN-based models except Photo dataset.
In addition, it has at least $1/20$ energy cost to the Riemannian baselines.

\begin{wrapfigure}{r}{0.35\textwidth}
\centering
\vspace{-0.2in}
\includegraphics[width=1\linewidth]{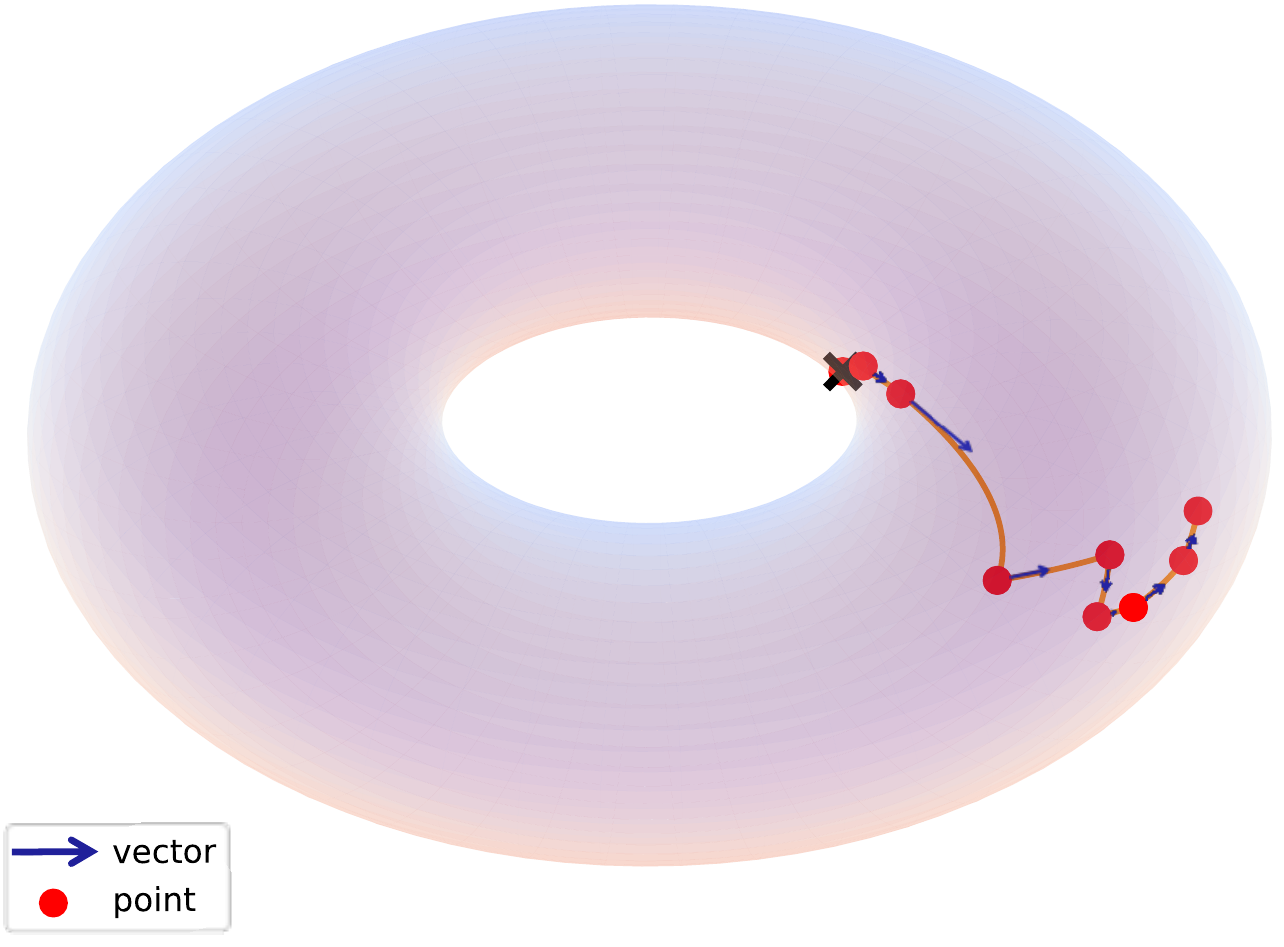}
\vspace{-0.25in}
\caption{Visualization on $\mathbb S^1 \times \mathbb S^1$}
\vspace{-0.3in}
\label{fig. visual}
\end{wrapfigure}

     \vspace{-0.12in}
\paragraph{Visualization \& Discussion.}
We empirically study the connection between the proposed \texttt{MSG} and manifold ODE.
In particular, we visualize a toy example of Zachary Karate Club dataset \cite{zachary1977information} on a $\mathbb S^1 \times \mathbb S^1$ in Fig. 5, 
where we plot each layer output on the manifold.
The red curve is the path connecting the layer input and layer output, and the blue one is the direction of the geodesic.
As shown in Fig. 5, the red and blue curves are coincided, that is, each layer solves an ODE describing the geodesic on the manifold.


\vspace{-0.15in}
\section{Conclusion}
\vspace{-0.1in}

In this paper, we study spiking GNN from a fundamentally different perspective of Riemannian geometry, and present a simple yet effective Manifold-valued Spiking GNN (\texttt{MSG}).
Concretely, we design a manifold spiking neuron which leverages the diffeomorphism to bridge spiking representations and manifold representations.
With the proposed neuron, we propose a new training algorithm with  Differentiation via Manifold, which no longer needs to recur the backward gradient and thus alleviates the high latency of previous methods.
An interesting theoretical result is that, \texttt{MSG} is essentially related to manifold ODE.
Extensive empirical results on benchmark datasets demonstrate the superior effectiveness and energy efficiency of the proposed \texttt{MSG}.

\vspace{-0.15in}
\section{Broader Impact and Limitations}
\label{impact}
 \vspace{-0.1in}
Our work brings together two previously separate domains: spiking neural network and Riemannian geometry, and presents a novel Manifold-valued Spiking GNN for energy-efficiency graph learning, especially for the large graphs. Our work is mainly a theoretical exploration, and not tied to particular applications. A positive societal impact is the possibility of decreasing carbon emissions in training large models. None of negative societal impacts we feel must be specifically highlighted here. 

 \vspace{-0.1in}
 \paragraph{Limitation.} 
 Our work as well as the previous spiking GNNs considers the undirected, homophilous graphs, while the spiking GNN on directed or heterophilous graphs still remains open.
 Also, readers may find it challenging to implement the proposed method. However, we provide downloadable code and will offer an easy-to-use interface.

\section*{Acknowledgement}
This work is supported in part by NSFC under grants 62202164 and 62322202.
Philip S. Yu is supported in part by NSF under grants III-2106758, and POSE-2346158.

\bibliography{neurips_2024}

\begin{thebibliography}{10}

\bibitem{hamilton2017inductive}
Hamilton, W., Z.~Ying, J.~Leskovec.
\newblock Inductive representation learning on large graphs.
\newblock \emph{Advances in neural information processing systems}, 30, 2017.

\bibitem{velickovic2018graph}
Veli{\v{c}}kovi{\'{c}}, P., G.~Cucurull, A.~Casanova, et~al.
\newblock {Graph Attention Networks}.
\newblock \emph{International Conference on Learning Representations}, 2018.
\newblock Accepted as poster.

\bibitem{chami2019hyperbolic}
Chami, I., Z.~Ying, C.~R{\'e}, et~al.
\newblock Hyperbolic graph convolutional neural networks.
\newblock In \emph{Advances in NeurIPS}, vol.~32. 2019.

\bibitem{xiong2022pseudo}
Xiong, B., S.~Zhu, N.~Potyka, et~al.
\newblock Pseudo-riemannian graph convolutional networks.
\newblock In \emph{Advances in NeurIPS}, vol.~32. 2022.

\bibitem{zhu2022spiking}
Zhu, Z., J.~Peng, J.~Li, et~al.
\newblock Spiking graph convolutional networks.
\newblock In \emph{Proceedings of the 31st IJCAI}, pages 2434--2440. ijcai.org,
  2022.

\bibitem{li2023graph}
Li, J., H.~Zhang, R.~Wu, et~al.
\newblock {A Graph is Worth 1-bit Spikes: When Graph Contrastive Learning Meets
  Spiking Neural Networks}.
\newblock \emph{arXiv preprint arXiv:2305.19306}, 2023.

\bibitem{maass1997networks}
Maass, W.
\newblock Networks of spiking neurons: The third generation of neural network
  models.
\newblock \emph{Neural networks}, 10(9):1659--1671, 1997.

\bibitem{brette2007simulation}
Brette, R., M.~Rudolph, T.~Carnevale, et~al.
\newblock Simulation of networks of spiking neurons: A review of tools and
  strategies.
\newblock \emph{Journal of computational neuroscience}, 23:349--398, 2007.

\bibitem{wang2022spiking}
Wang, B., B.~Jiang.
\newblock {Spiking GATs}: Learning graph attentions via spiking neural network.
\newblock \emph{arXiv preprint arXiv:2209.13539}, 2022.

\bibitem{yang2022spiking}
Yang, H., R.~Zhang, Q.~Kong, et~al.
\newblock Spiking variational graph auto-encoders for efficient graph
  representation learning.
\newblock \emph{arXiv preprint arXiv:2211.01952}, 2022.

\bibitem{yin2024continuous}
Yin, N., M.~Wang, L.~Shen, et~al.
\newblock Continuous spiking graph neural networks.
\newblock \emph{CoRR}, abs/2404.01897, 2024.

\bibitem{sarkar2012low}
Sarkar, R.
\newblock Low distortion delaunay embedding of trees in hyperbolic plane.
\newblock In \emph{Proceedings of the 19th International Symposium on Graph
  Drawing}, vol. 7034 of \emph{Lecture Notes in Computer Science}, pages
  355--366. Springer, 2011.

\bibitem{bachmann2020constanta}
Bachmann, G., G.~B{\'{e}}cigneul, O.~Ganea.
\newblock Constant curvature graph convolutional networks.
\newblock In \emph{Proceedings of the 37th ICML}, vol. 119, pages 486--496.
  {PMLR}, 2020.

\bibitem{wu2021training}
Wu, H., Y.~Zhang, W.~Weng, et~al.
\newblock Training {{Spiking Neural Networks}} with {{Accumulated Spiking
  Flow}}.
\newblock \emph{Proceedings of the AAAI Conference on Artificial Intelligence},
  35(12):10320--10328, 2021.

\bibitem{meng2022training}
Meng, Q., M.~Xiao, S.~Yan, et~al.
\newblock Training high-performance low-latency spiking neural networks by
  differentiation on spike representation.
\newblock In \emph{Proceedings of CVPR}, pages 12434--12443. {IEEE}, 2022.

\bibitem{wu2023tandem}
Wu, J., Y.~Chua, M.~Zhang, et~al.
\newblock A {{Tandem Learning Rule}} for {{Effective Training}} and {{Rapid
  Inference}} of {{Deep Spiking Neural Networks}}.
\newblock \emph{IEEE Transactions on Neural Networks and Learning Systems},
  34(1):446--460, 2023.

\bibitem{lou2020neural}
Lou, A., D.~Lim, I.~Katsman, et~al.
\newblock Neural manifold ordinary differential equations.
\newblock In \emph{Advances in NeurIPS}. 2020.

\bibitem{kipf2016semi}
Kipf, T.~N., M.~Welling.
\newblock Semi-supervised classification with graph convolutional networks.
\newblock In \emph{Proceedings of the 5th ICLR}. OpenReview.net, 2017.

\bibitem{wu2019simplifying}
Wu, F., A.~Souza, T.~Zhang, et~al.
\newblock Simplifying graph convolutional networks.
\newblock In \emph{International conference on machine learning}, pages
  6861--6871. {PMLR}, 2019.

\bibitem{feng2019hypergrapha}
Feng, Y., H.~You, Z.~Zhang, et~al.
\newblock Hypergraph {{Neural Networks}}.
\newblock \emph{Proceedings of the AAAI Conference on Artificial Intelligence},
  33(01):3558--3565, 2019.

\bibitem{guo2021hierarchicala}
Guo, K., Y.~Hu, Y.~Sun, et~al.
\newblock Hierarchical {{Graph Convolution Network}} for {{Traffic
  Forecasting}}.
\newblock \emph{Proceedings of the AAAI Conference on Artificial Intelligence},
  35(1):151--159, 2021.

\bibitem{liu2019hyperbolica}
Liu, Q., M.~Nickel, D.~Kiela.
\newblock Hyperbolic {{Graph Neural Networks}}.
\newblock In \emph{Advances in {{Neural Information Processing Systems}}},
  vol.~32. Curran Associates, Inc., 2019.

\bibitem{coors2018spherenet}
Coors, B., A.~P. Condurache, A.~Geiger.
\newblock {SphereNet}: Learning spherical representations for detection and
  classification in omnidirectional images.
\newblock In \emph{Proceedings of ECCV}, pages 518--533. 2018.

\bibitem{gu2018learning}
Gu, A., F.~Sala, B.~Gunel, et~al.
\newblock Learning mixed-curvature representations in product spaces.
\newblock In \emph{Proceedings of the 7th ICLR}. OpenReview.net, 2019.

\bibitem{zhang2021switch}
Zhang, S., Y.~Tay, W.~Jiang, et~al.
\newblock {Switch Spaces}: Learning product spaces with sparse gating.
\newblock \emph{CoRR}, abs/2102.08688, 2021.

\bibitem{aaai22SelfMix}
Sun, L., Z.~Zhang, J.~Ye, et~al.
\newblock A self-supervised mixed-curvature graph neural network.
\newblock In \emph{Proceedings of the 36th AAAI}, pages 4146--4155. 2022.

\bibitem{SunL24AAAI}
Sun, L., Z.~Huang, Z.~Wang, et~al.
\newblock Motif-aware riemannian graph neural network with
  generative-contrastive learning.
\newblock In \emph{Proceedings of the 38th AAAI}, pages 9044--9052. 2024.

\bibitem{law2021ultrahyperbolic}
Law, M.
\newblock Ultrahyperbolic neural networks.
\newblock In \emph{Advances in NeurIPS}. 2021.

\bibitem{gao2020learning}
Gao, Z., Y.~Wu, Y.~Jia, et~al.
\newblock Learning to optimize on {SPD} manifolds.
\newblock In \emph{Proceedings of CVPR}, pages 7697--7706. Computer Vision
  Foundation / IEEE, 2020.

\bibitem{dong2017deep}
Dong, Z., S.~Jia, C.~Zhang, et~al.
\newblock Deep manifold learning of symmetric positive definite matrices with
  application to face recognition.
\newblock In \emph{Proceedings of the AAAI Conference on Artificial
  Intelligence}, vol.~31. 2017.

\bibitem{SunL24ICML}
Sun, L., Z.~Huang, H.~Peng, et~al.
\newblock Lsenet: Lorentz structural entropy neural network for deep graph
  clustering.
\newblock In \emph{Proceedings of the 41st ICML}. OpenReview.net, 2024.

\bibitem{SunL24WWW}
Sun, L., J.~Hu, S.~Zhou, et~al.
\newblock Riccinet: Deep clustering via a riemannian generative model.
\newblock In \emph{Proceedings of the ACM Web Conference 2024 (WWW’24)},
  pages 4071--4082. 2024.

\bibitem{SunL23ICDM}
Sun, L., Z.~Huang, H.~Wu, et~al.
\newblock Deepricci: Self-supervised graph structure-feature co-refinement for
  alleviating over-squashing.
\newblock In \emph{Proceedings of the 23rd ICDM}, pages 558--567. 2023.

\bibitem{SunL23AAAI}
Sun, L., J.~Ye, H.~Peng, et~al.
\newblock Self-supervised continual graph learning in adaptive riemannian
  spaces.
\newblock In \emph{Proceedings of the 37th AAAI}, pages 4633--4642. 2023.

\bibitem{HVGNN}
Sun, L., Z.~Zhang, J.~Zhang, et~al.
\newblock Hyperbolic variational graph neural network for modeling dynamic
  graphs.
\newblock In \emph{Proceedings of the 35th AAAI}, pages 4375--4383. 2021.

\bibitem{SunL22CIKM}
Sun, L., J.~Ye, H.~Peng, et~al.
\newblock A self-supervised riemannian gnn with time varying curvature for
  temporal graph learning.
\newblock In \emph{Proceedings of the 31st ACM CIKM}, pages 1827--1836. 2022.

\bibitem{SunL24SIGIR}
Sun, L., J.~Hu, M.~Li, et~al.
\newblock R-ode: Ricci curvature tells when you will be informed.
\newblock In \emph{Proceedings of the ACM SIGIR}. 2024.

\bibitem{cao2015spiking}
Cao, Y., Y.~Chen, D.~Khosla.
\newblock Spiking deep convolutional neural networks for energy-efficient
  object recognition.
\newblock \emph{International Journal of Computer Vision}, 113:54--66, 2015.

\bibitem{cao2024spiking}
Cao, J., Z.~Wang, H.~Guo, et~al.
\newblock Spiking denoising diffusion probabilistic models.
\newblock In \emph{Proceedings of the IEEE/CVF Winter Conference on
  Applications of Computer Vision}, pages 4912--4921. 2024.

\bibitem{xu2021exploiting}
Xu, M., Y.~Wu, L.~Deng, et~al.
\newblock Exploiting spiking dynamics with spatial-temporal feature
  normalization in graph learning.
\newblock In \emph{Proceedings of the 30th IJCAI}, pages 3207--3213. ijcai.org,
  2021.

\bibitem{poli2019graph}
Poli, M., S.~Massaroli, J.~Park, et~al.
\newblock Graph neural ordinary differential equations.
\newblock \emph{arXiv preprint arXiv:1911.07532}, 2019.

\bibitem{li2023scaling}
Li, J., Z.~Yu, Z.~Zhu, et~al.
\newblock Scaling up dynamic graph representation learning via spiking neural
  networks.
\newblock In \emph{Proceedings of the 37th AAAI}, vol.~37, pages 8588--8596.
  2023.

\bibitem{zhao2024dynamic}
Zhao, H., X.~Yang, C.~Deng, et~al.
\newblock Dynamic reactive spiking graph neural network.
\newblock In \emph{Proceedings of the 38th AAAI}, vol.~38, pages 16970--16978.
  2024.

\bibitem{yin2024dynamic}
Yin, N., M.~Wang, Z.~Chen, et~al.
\newblock Dynamic spiking graph neural networks.
\newblock In \emph{Proceedings of the 38th AAAI}, vol.~38, pages 16495--16503.
  2024.

\bibitem{salinas2002integrate}
Salinas, E., T.~J. Sejnowski.
\newblock Integrate-and-fire neurons driven by correlated stochastic input.
\newblock \emph{Neural computation}, 14(9):2111--2155, 2002.

\bibitem{burkitt2006review}
Burkitt, A.~N.
\newblock A review of the integrate-and-fire neuron model: I. homogeneous
  synaptic input.
\newblock \emph{Biological Cybernetics}, 95(1):1--19, 2006.

\bibitem{huh2018gradient}
Huh, D., T.~J. Sejnowski.
\newblock Gradient {{Descent}} for {{Spiking Neural Networks}}.
\newblock In \emph{Advances in {{Neural Information Processing Systems}}},
  vol.~31. 2018.

\bibitem{neftci2019surrogate}
Neftci, E.~O., H.~Mostafa, F.~Zenke.
\newblock Surrogate gradient learning in spiking neural networks: Bringing the
  power of gradient-based optimization to spiking neural networks.
\newblock \emph{{IEEE} Signal Process. Mag.}, 36(6):51--63, 2019.

\bibitem{li2021differentiable}
Li, Y., Y.~Guo, S.~Zhang, et~al.
\newblock Differentiable {{Spike}}: {{Rethinking Gradient-Descent}} for
  {{Training Spiking Neural Networks}}.
\newblock In \emph{Advances in {{Neural Information Processing Systems}}},
  vol.~34. 2021.

\bibitem{shchur2018pitfalls}
Shchur, O., M.~Mumme, A.~Bojchevski, et~al.
\newblock Pitfalls of graph neural network evaluation.
\newblock \emph{arXiv preprint arXiv:1811.05868}, 2018.

\bibitem{chen2021fully}
Chen, W., X.~Han, Y.~Lin, et~al.
\newblock Fully hyperbolic neural networks.
\newblock In \emph{Proceedings of the 60th ACL}, pages 5672--5686. Association
  for Computational Linguistics, 2022.

\bibitem{sun2024spikegraphormer}
Sun, Y., D.~Zhu, Y.~Wang, et~al.
\newblock {SpikeGraphormer}: A high-performance graph transformer with spiking
  graph attention.
\newblock \emph{arXiv preprint arXiv:2403.15480}, 2024.

\bibitem{geoopt19}
B{\'{e}}cigneul, G., O.~Ganea.
\newblock Riemannian adaptive optimization methods.
\newblock In \emph{Proceedings of ICLR}. OpenReview.net, 2019.

\bibitem{paszke2019pytorcha}
Paszke, A., S.~Gross, F.~Massa, et~al.
\newblock {{PyTorch}}: {{An Imperative Style}}, {{High-Performance Deep
  Learning Library}}.
\newblock In \emph{Advances in {{Neural Information Processing Systems}}},
  vol.~32. 2019.

\bibitem{zachary1977information}
Zachary, W.~W.
\newblock An information flow model for conflict and fission in small groups.
\newblock \emph{Journal of anthropological research}, 33(4):452--473, 1977.

\bibitem{tu2011introduction}
Tu, L.~W.
\newblock \emph{An {{Introduction}} to {{Manifolds}}}.
\newblock Springer New York, 2011.

\bibitem{crouch1993numerical}
Crouch, P.~E., R.~Grossman.
\newblock Numerical integration of ordinary differential equations on
  manifolds.
\newblock \emph{Journal of Nonlinear Science}, 3(1):1--33, 1993.

\bibitem{Petersen16}
Petersen, P.
\newblock \emph{Riemannian Geometry, 3rd edition}.
\newblock Springer-Verlag, 2016.

\bibitem{zhou2022spikformer}
Zhou, Z., Y.~Zhu, C.~He, et~al.
\newblock Spikformer: When spiking neural network meets transformer.
\newblock \emph{arXiv preprint arXiv:2209.15425}, 2022.

\end{thebibliography}
\bibliographystyle{neurips_2024}



\newpage
\appendix

 The appendix is organized as follows: 
 \begin{enumerate}
\item \textbf{Notation table},
\item \textbf{Proofs} of Theorem 4.1 and Theorem 5.2, 
\item \textbf{Derivation} of the Jacobian, 
\item \textbf{Riemannian geometry} including geodesically complete manifold, stereographic projection and $\kappa$-stereographic model, and Cartesian product and product manifold, 
\item \textbf{Empirical details}, i.e., datasets/baselines description, theoretical  energy consumption and implementation notes,
\item \textbf{Additional results} of link prediction, layer-wise gradient and the visualization.
 \end{enumerate}

\section{Notations}
\label{append. notation}
We summarize the important notations of our paper in Table \ref{table. notation}.
\begin{table}[htb]
\centering
\caption{Notations.}
\label{table. notation}
\begin{tabular}{|c|c|c|}
\hline
\textbf{Notation} & \textbf{Description}   \\
\hline
$\mathcal{M},\mathcal{N}$        & Smooth manifolds         \\
\hline
 $\mathbf{x}, \mathbf{y}, \mathbf{z}$  &  Points on manifolds \\
 \hline
 $\mathbf{o}$ & The original point on manifold \\
 \hline
 $T_\mathbf{z}\mathcal{M}$ & The tangent space at point $\mathbf{z}$ \\
  \hline
 $T\mathcal{M}$ & The tangent bundle of the manifold $\mathcal M$ \\
  \hline
 $u$ & The vector field over the manifold, described by an ODE \\
 \hline
 $T^*_\mathbf{z}\mathcal{N}$ & The dual space of $T_\mathbf{z}\mathcal{N}$ \\
 \hline
 $D_\mathbf{z}f$ & The differential of $f$ at point $\mathbf{z}$ \\
 \hline
 $(df)_\mathbf{z}$ & The differential $1-$form of $f$ at point $\mathbf{z}$ \\
 \hline
 $\{(\partial/\partial z^1)|_\mathbf{z},...,(\partial/\partial z^n)|_\mathbf{z}\}$ & A basis of the tangent space $T_\mathbf{z}\mathcal{N}$ \\
 \hline
 $\{ (dz^1)_\mathbf{z}, ..., (dz^n)_\mathbf{z} \}$ & A basis of $T^*_\mathbf{z}\mathcal{N}$\\
 \hline
 $\nabla \mathcal{L}$ & The gradient of a smooth scalar function $\mathcal{L}$ \\
 \hline
 $\operatorname{Exp}_\mathbf{z}(\cdot)$ & The exponential map at point $\mathbf{z}$ \\
 \hline
 $\operatorname{Log}_\mathbf{z}(\cdot)$ & The logarithmic map at point $\mathbf{z}$ \\
 \hline
 $\mathbb{S}^d$ & $d$-dimensional Sphere model of hyperspherical space \\
 \hline
  $\mathbb{E}^d$ & $d$-dimensional Euclidean space \\
 \hline
 $\mathbb{H}^d$ & $d$-dimensional Lorentz model of Hyperbolic space \\
 \hline
 $\mathcal{V}, \mathcal{E}, \mathbf{F}, \mathbf{A}$ & Node set $\mathcal{V}$, edge set $\mathcal{E}$, feature matrix $\mathbf{F}$ and adjacency matrix $\mathbf{A}$\\
  \hline
 $\mathcal{G}=(\mathcal{V}, \mathcal{E}, \mathbf{F}, \mathbf{A})$ & A graph defined on $\mathcal{V}$, $\mathcal{E}$, $\mathbf{F}$ and  $\mathbf{A}$\\
 \hline
 $\Omega_i$ & Neighbourhood of node $i$ \\
   \hline
 $\mathcal{F}_\theta$ & A graph encoder with parameters $\theta$ \\
 \hline
 $V[t]$ & Membrane potential of a spiking neuron at time step $t$ \\
 \hline
 $H(\cdot)$ & The Heaviside step function \\
 \hline
 $S[t]$ & Spikes fired by a spiking neuron at time step $t$ \\
 \hline
 $V_{th}$ & Threshold membrane potential of a spiking neuron \\
 \hline
  $l$ & The index of spiking layer in neural network\\
 \hline
 $\operatorname{MSNeuron}$ & The proposed manifold spiking neuron \\
 \hline
\end{tabular}
 \vspace{-0.2in}
\end{table}

\section{Proofs}
\label{proof}
 \vspace{-0.05in}
 
In this section, we demonstrate the proofs of Theorem \ref{grad} (Backward Gradient) and Theorem \ref{ode} (\texttt{MSG} as Dynamic Chart Solver).

 \vspace{-0.05in}
\subsection{Proof of Proposition \ref{grad}}
\label{proof. grad}

First, we give the formal definition of the pullback.
Given a smooth map $\phi: \mathcal M \to \mathcal N$ connecting two manifolds $\mathcal M$ and $\mathcal N$, 
and a real, smooth function $f : \mathcal N \to \mathbb R$, 
the pullback of $f $ by $\phi$ is the smooth function $\phi^*f $ on $\mathcal M$ defined by $(\phi^*f)(\mathbf x)=f(\phi(\mathbf x))$.

Next, we introduce some properties of the pullback in the smooth manifold \cite{tu2011introduction} 
(i.e., differential 1-form of a smooth function, communication, and pullback of a sum and a product), 
supporting the derivation of the backward gradient (Theorem 4.1).
\begin{lemma}[Differential 1-form of a smooth function]
\label{lemma. form}
    For a point $\mathbf{z} \in \mathcal{N}$ related with a coordinate chart $(U, z^1, ..., z^n)$, 
    there is a series of covectors $\{ (dz^1)_\mathbf{z}, ..., (dz^n)_\mathbf{z} \}$ forming a basis of $T^*_\mathbf{z}\mathcal{N}$ 
    dual to the basis $\{(\partial/\partial z^1)|_\mathbf{z},...,(\partial/\partial z^n)|_\mathbf{z}\}$ of tangent space $T_\mathbf{z}\mathcal{N}$. 
    Then, for any smooth function $f$ on $\mathcal{N}$ restrict to $U$, the differential 1-form of $f$ is
    \begin{align}
        df = \sum_{i=1}^n\frac{\partial f}{\partial z^i}dz^i.
    \end{align}
\end{lemma}
\begin{lemma}[Communication]
\label{lemma. commu}
    Let $F: \mathcal{N}\rightarrow\mathcal{M}$ be a smooth map, 
    for any smooth function $g$ on $\mathcal{M}$, we have $F^*(dg)=d(F^*g)$.
\end{lemma}
\begin{lemma}[Pullback of a sum and a product]
\label{lemma. sum}
    Let $F: \mathcal{N}\rightarrow\mathcal{M}$ be a smooth map, 
    $g$ is a smooth scalar function on $\mathcal{M}$, 
    and $\omega, \gamma$ are differential 1-forms on $\mathcal{M}$. 
    Then, we have
    \begin{align}
        F^*(\omega + \gamma) = F^*\omega + F^*\gamma \\
        F^*(g\omega) = (F^*g)(F^*\omega).
    \end{align}
\end{lemma}


Given the properties of the pullback, we derive the closed-form backward gradient of the real function on the manifold, and prove Theorem 4.1.

\textbf{Theorem 4.1} (Backward Gradient) 
\emph{Let $\mathcal{L}$ be the scalar-valued function, 
and $\mathbf{z}^l$ is the output of $l$-th layer with parameter $\mathbf{W}^l$, 
which is delivered by tangent vector $\mathbf{v}^l$.
 Then, the gradient of function $\mathcal{L}$ w.r.t. $\mathbf{W}^l$ is given as follows:
\begin{align}
    \nabla_{\mathbf{W}^l} \mathcal{L} = [\frac{\partial \mathbf{v}^{l-1}}{\partial \mathbf{W}^l}]^* [D_{\mathbf{v}^{l-1}}\phi^{l-1}]^* \nabla_{\mathbf{z}^l} \mathcal{L},    \quad
     \nabla_{\mathbf{z}^l} \mathcal{L} = [D_{\mathbf{z}^{l}}\psi^{l}]^* \nabla_{\mathbf{z}^{l+1}} \mathcal{L}, 
\end{align}
where $\phi^{l-1}(\cdot)=\operatorname{Exp}_{\mathbf{z}^{l-1}}(\cdot)$, $\psi^{l}(\cdot)=\operatorname{Exp}_{(\cdot)}({\mathbf{v}^{l}})$, 
and $[\cdot]^*$ means the matrix form of pullback.
}
\begin{proof}
    Given $\mathbf{z}^l, \mathbf{z}^{l+1}$ in $\mathcal{M}$, and $F: \mathcal{M} \rightarrow \mathcal{M}$ be the smooth map such that $F(\mathbf{z}^l)=\mathbf{z}^{l+1}$. Consider scalar loss function $L: \mathcal{M}\rightarrow \mathbb{R}$, if we relate $\mathbf{z}^l$ with a chart $(U, x^1,...,x^m)$ and $\mathbf{z}^{l+1}$ with $(V, y^1, ..., y^m)$, the gradients of $L$ at $\mathbf{z}^{l+1}$ and $\mathbf{z}^{l}$ are given by Lemma. \ref{lemma. form},
    \begin{align}
        \nabla_{\mathbf{z}^{l+1}} \mathcal{L} = \sum_{i}\frac{\partial \mathcal{L}}{\partial y^i}\bigg|_{\mathbf{z}^{l+1}} dy^i \\
        \nabla_{\mathbf{z}^{l}} (\mathcal{L}\circ F) = \sum_{i}\frac{\partial \mathcal{L}\circ F}{\partial x^i}\bigg|_{\mathbf{z}^{l}} dx^i.
    \end{align}
    Then, we apply the pullback $F^*$ on $\nabla_{\mathbf{z}^{l+1}} L$ that 
    \begin{align}
        F^*(\nabla_{\mathbf{z}^{l+1}} \mathcal{L}) &= F^*\sum_{i}\frac{\partial \mathcal{L}}{\partial y^i}\bigg|_{\mathbf{z}^{l+1}} dy^i|_{\mathbf{z}^{l+1}} \\
        &= \sum_{i}(F^*\frac{\partial \mathcal{L}}{\partial y^i}\bigg|_{\mathbf{z}^{l+1}}) (F^*dy^i|_{\mathbf{z}^{l+1}}) \quad \text{from } \textbf{ Lemma. \ref{lemma. sum}} \\
        &= \sum_{i}(\frac{\partial \mathcal{L}}{\partial y^i} \circ F)|_{\mathbf{z}^{l}} (d(F^*y^i)|_{\mathbf{z}^{l}}) \quad \text{from } \textbf{ Lemma. \ref{lemma. commu}}\\
        &= \sum_{i}(\frac{\partial \mathcal{L}}{\partial y^i} \circ F)|_{\mathbf{z}^{l}} (d(y^i \circ F))|_{\mathbf{z}^{l}} \\
        &= \sum_{i}(\frac{\partial \mathcal{L}}{\partial y^i}\bigg|_{\mathbf{z}^{l+1}}) (\sum_j\frac{\partial F^i}{\partial x^j}dx^j)|_{\mathbf{z}^{l}} \quad \text{from }  \textbf{ Lemma. \ref{lemma. form}}\\
        &=\sum_{i,j} \frac{\partial \mathcal{\mathcal{L}}}{\partial y^i}\bigg|_{\mathbf{z}^{l+1}} \frac{\partial F^i}{\partial x^j}\bigg|_{\mathbf{z}^{l}}dx^j|_{\mathbf{z}^{l}},
    \end{align}
    Then, we can find that the matrix form of the pullback $F^*$ can be written as the transpose of the Jacobian matrix of $F$, denoted as $[\frac{\partial F^i}{\partial x^j}|_{\mathbf{z}^{l}}]^*$ or $[D_{\mathbf{z}^l}F]^*$.
    The derivation of $\nabla_{\mathbf{W}^l} \mathcal{L}$ is similar, we only need to use an addition process like above on $\nabla_{\mathbf{v}^{l-1}} \mathcal{L}$.
\end{proof}

Note that, we give the closed-form expression of exponential map, logarithmic map, and parallel transport for hyperbolic and hyperspherical space,  and derive the corresponding Jacobian in Sec. \ref{formulas}.

     \vspace{-0.05in}
\subsection{Proof of Theorem \ref{ode}}
\label{proof. ode}
\textbf{Theorem 5.2} (\texttt{MSG} as Dynamic Chart Solver)
\emph{If $\mathbf{y}(t): [\tau, \tau + \epsilon] \to \mathbb R^n$ is the solution of 
         \vspace{-0.03in}
   \begin{align}
    \label{eq. Eode}
        \frac{d\mathbf{y}(t)}{dt} = (D_{\operatorname{Exp}_\mathbf{z}(\mathbf{y(t)})} \operatorname{Log}_\mathbf{z}) u(\operatorname{Exp}_\mathbf{z}(\mathbf{y}(t)), t),
            \vspace{-0.05in}
    \end{align}
    then 
  $\mathbf{z}(t)=\operatorname{Exp}_\mathbf{z}(\mathbf{y}(t))$ 
  is a valid solution to the manifold ODE of Eq. (\ref{mode}) on $t \in [\tau, \tau + \epsilon]$, where $\mathbf{z}=\mathbf{z}(\tau)$.
 If $\mathbf{y}(t)$ is given by the first-order approximation with the $\epsilon$ small enough,
  \vspace{-0.03in}
    \begin{align}
        \mathbf{y}(\tau + \epsilon)=\epsilon \cdot(D_\mathbf{z}\operatorname{Log}_\mathbf{z})u(\mathbf{z}(\tau), \tau),
            \vspace{-0.05in}
    \end{align}
    then the update process of  Eqs. (\ref{eq. x}) and  (\ref{eq. z}) in \texttt{MSG} is equivalent to Dynamic Chart Solver in Eq. (\ref{dynamicChartSolver}).}
\begin{proof}
    Let $t \in [\tau, \tau + \epsilon]$, then we have
    \begin{align}
        \frac{d\mathbf{z}(t)}{dt} &= (D_{\mathbf{y(t)}}\operatorname{Exp}_\mathbf{z})\frac{d\mathbf{y}(t)}{dt} \\
        &= (D_{\mathbf{y(t)}}\operatorname{Exp}_\mathbf{z}) (D_{\operatorname{Exp}_\mathbf{z}(\mathbf{y(t)})} \operatorname{Log}_\mathbf{z}) u(\operatorname{Exp}_\mathbf{z}(\mathbf{y}(t)), t) \\
        &= (D_{\mathbf{y(t)}}\operatorname{Exp}_\mathbf{z}) (D_{\mathbf{z}(t)} \operatorname{Log}_\mathbf{z}) u(\operatorname{Exp}_\mathbf{z}(\mathbf{y}(t)), t) \\
        &= u(\mathbf{z}(t), t).
    \end{align}
    Consider two adjacent charts $(U_1, \operatorname{Log}_{\mathbf{z}_1})$ and $(U_2, \operatorname{Log}_{\mathbf{z}_2})$, 
    such that $\mathbf{z}_1\in U_1$ and $\mathbf{z}_2 \in U_1 \cap U_2$. 
    Note that, in interval $[\tau, \tau+\epsilon]$, $\mathbf{z}(\tau)=\mathbf{z}_1$ and $\mathbf{z}(\tau + \epsilon)=\mathbf{z}_2$, 
    we have $\mathbf{y}(\tau)=\operatorname{Exp}_{\mathbf{z}_1}(\mathbf{z}_1)=\mathbf{0}$. 
    With the first-order approximation, 
    $\mathbf{y}(\tau + \epsilon)$ is thus given by
    \begin{align}
        \mathbf{y}(\tau + \epsilon) &= \mathbf{y}(\tau) + \epsilon \cdot(D_{\operatorname{Exp}_{\mathbf{z}_1}(\mathbf{y(\tau)})} \operatorname{Log}_\mathbf{{z}_1}) u(\operatorname{Exp}_\mathbf{z}(\mathbf{y}(\tau)), \tau) \\
        &=\epsilon \cdot(D_{\mathbf{z}_1}\operatorname{Log}_{\mathbf{z}_1})u(\mathbf{z}(\tau), \tau)
        \label{eq. approx}
    \end{align}
    Also, Eq. (\ref{eq. approx}) can be treated as a step in the Euler solver \cite{crouch1993numerical} for a small $\epsilon$.
    Finally, we have $\mathbf{z}(\tau + \epsilon)=\operatorname{Exp}_{\mathbf{z}_1}(\mathbf{y}(\tau + \epsilon))=\mathbf{z}_2$, 
   ending the process of dynamic chart solver. 
   That is, \texttt{MSG} considers the logarithmic map to define the charts, and is equivalent to Dynamic Chart Solver (Definition \ref{def. dynamic}), completing the proof.
\end{proof}

     \vspace{-0.1in}
\section{Deviation of Jacobian} \label{formulas}
     \vspace{-0.05in}

We instantiate the proposed \texttt{MSG} in the Lorentz model $\mathbb H$ of hyperbolic space, sphere model $\mathbb S$ of hyperspherical space, and the products of $\mathbb H$'s or/and $\mathbb S$'s.
Accordingly, we derive the Jacobian in $\mathbb H$ and $\mathbb S$, and introduce the construction in the products in \ref{product}.

     \vspace{-0.05in}
\subsection{Hyperbolic Space}

     \vspace{-0.05in}
\paragraph{Lorentz Model}
The $d$-dimensional Lorentz model  $\mathbb{H}^d$
 is defined on the $(d+1)$-dimensional manifold of
 $\{\mathbf{z}=[z_0, z_1, \cdots, z_d]^T \in \mathbb{R}^{d+1} | \langle \mathbf{z}, \mathbf{z} \rangle_\mathcal{L}=-1, z_0 > 0\}$ 
 \footnote{We utilize the manifold of standard curvature for model instantiation, i.e., constant curvature of $-1$ for hyperbolic space, and $1$ for hyperspherical space. Note that, hyperbolic/hyperspherical  spaces of different constant curvatures are mathematically equivalent in essence.},
 equipped with the Minkowski inner product, 
 \vspace{-0.1in}
\begin{align}
    \langle \mathbf{u}, \mathbf{v} \rangle_\mathcal{L} = -u_0v_0 + \sum_{i=1}^d u_iv_i.
     \vspace{-0.05in}
\end{align}
The tangent space at point $\mathbf{z} \in \mathbb{H}^d$ is $T_\mathbf{z}\mathbb{H}^d=\{\mathbf{v} \in \mathbb{R}^{d+1}  | \langle \mathbf{z}, \mathbf{v} \rangle_\mathcal{L}=0\}$,
and $\operatorname{Proj}_\mathbf{z}(\mathbf{u}) = \mathbf{u} + \langle \mathbf{z}, \mathbf{u} \rangle_\mathcal{L}\mathbf{z}$  is to project a vector $u \in \mathbb{R}^{d+1}$ into the tangent space $T_\mathbf{z}\mathbb{H}^d$.
The Lorentz norm of tangent vector is defined as $\lVert \mathbf{v} \rVert_\mathcal{L}=\sqrt{\langle \mathbf{z}, \mathbf{z} \rangle_\mathcal{L}}$.

Theorem \ref{grad} requires the Jocabian of $\phi(\cdot)=\operatorname{Exp}_{\mathbf{z}}(\cdot)$ and $\psi(\cdot)=\operatorname{Exp}_{(\cdot)}(\mathbf{v})$,
and  Lorentz model has the closed-form exponential map given as follows,
\begin{align}
    \operatorname{Exp}_\mathbf{z}(\mathbf{v}) = \cosh({\lVert \mathbf{v} \rVert_\mathcal{L}})\mathbf{z} + \frac{\sinh(\lVert \mathbf{v} \rVert_\mathcal{L})}{\lVert \mathbf{v} \rVert_\mathcal{L}}\mathbf{v}.
    \label{lorentz-exp}
\end{align}
The inverse of exponential map (i.e,  the logarithmic map) is
\begin{align}
    \operatorname{Log}_\mathbf{z}(\mathbf{x}) = \frac{\operatorname{arcosh}(\langle \mathbf{z}, \mathbf{x} \rangle_\mathcal{L})}{\sinh(\operatorname{arcosh}(\langle \mathbf{z}, \mathbf{x} \rangle_\mathcal{L}))}(\mathbf{x} - \langle \mathbf{z}, \mathbf{x} \rangle_\mathcal{L}\mathbf{z})
\end{align}

\paragraph{Deviation  of Jacobian}

We first calculate the Jacobian of $\psi$, and it is given as
\begin{align}
    D_\mathbf{z}\psi = \cosh({\lVert \mathbf{v} \rVert_\mathcal{L}}) \mathbf{I}.
\end{align}
Note that, Jacobian of $\phi$ needs the Jacobian of $\lVert \mathbf{v} \rVert_\mathcal{L}$, which is derived as
\begin{align}
    D_\mathbf{v}\lVert \mathbf{v} \rVert_\mathcal{L} = \frac{d}{d\langle \mathbf{v}, \mathbf{v} \rangle_\mathcal{L}}(\sqrt{\langle \mathbf{v}, \mathbf{v} \rangle}_\mathcal{L})D_\mathbf{v}(\langle \mathbf{v}, \mathbf{v} \rangle_\mathcal{L})
    = \frac{1}{\lVert \mathbf{v} \rVert_\mathcal{L}}\hat{\mathbf{v}}^T,
\end{align}
where $\hat{\mathbf{v}} = [-v_0, v_1, ..., v_d]^T$.
Then, we compute the derivative of the first term of  Eq. \ref{lorentz-exp}.
\begin{align}
    D_\mathbf{v}\cosh({\lVert \mathbf{v} \rVert_\mathcal{L}})\mathbf{z} = \frac{d}{d\lVert \mathbf{v} \rVert_\mathcal{L}}(\cosh({\lVert \mathbf{v} \rVert_\mathcal{L}}))\mathbf{z}(D_\mathbf{v}\lVert \mathbf{v} \rVert_\mathcal{L})
&=\frac{\sinh(\lVert \mathbf{v} \rVert_\mathcal{L})}{\lVert \mathbf{v} \rVert_\mathcal{L}}\mathbf{z}\hat{\mathbf{v}}^T,
\end{align}
and the derivative of the second term is derived as
\begin{align}
    D_\mathbf{v}\frac{\sinh(\lVert \mathbf{v} \rVert_\mathcal{L})}{\lVert \mathbf{v} \rVert_\mathcal{L}}\mathbf{v} &= D_\mathbf{v}(\frac{\sinh(\lVert \mathbf{v} \rVert_\mathcal{L})}{\lVert \mathbf{v} \rVert_\mathcal{L}})\mathbf{v} + \frac{\sinh(\lVert \mathbf{v} \rVert_\mathcal{L})}{\lVert \mathbf{v} \rVert_\mathcal{L}}D_\mathbf{v}\mathbf{v} \\
    &=\frac{\lVert \mathbf{v} \rVert_\mathcal{L}\cosh(\lVert \mathbf{v} \rVert_\mathcal{L}) - \sinh(\lVert \mathbf{v} \rVert_\mathcal{L})}{\lVert \mathbf{v} \rVert_\mathcal{L}^3}\mathbf{v}\hat{\mathbf{v}}^T + \frac{\sinh(\lVert \mathbf{v} \rVert_\mathcal{L})}{\lVert \mathbf{v} \rVert_\mathcal{L}}\mathbf{I}
\end{align}
Summing up above equations, we finally have
\begin{align}
    D_\mathbf{v}\phi = \frac{\lVert \mathbf{v} \rVert_\mathcal{L}\cosh(\lVert \mathbf{v} \rVert_\mathcal{L}) - \sinh(\lVert \mathbf{v} \rVert_\mathcal{L})}{\lVert \mathbf{v} \rVert_\mathcal{L}^3}\mathbf{v}\hat{\mathbf{v}}^T + \frac{\sinh(\lVert \mathbf{v} \rVert_\mathcal{L})}{\lVert \mathbf{v} \rVert_\mathcal{L}}(\mathbf{I} + \mathbf{z}\hat{\mathbf{v}}^T),
\end{align}
where $\mathbf I$ is the identity matrix.

\subsection{Hyperspherical Space}

\paragraph{Sphere Model}
The sphere model $\mathbb{S}^d$  is defined on the $(d+1)$-dimensional manifold of 
$\{\mathbf{z}=[z_0, z_1, \cdots, z_d]^T \in \mathbb{R}^{d+1} | \langle \mathbf{z}, \mathbf{z} \rangle=1, z_0 > 0\}$ 
 with the standard inner product $\langle \mathbf{x}, \mathbf{y} \rangle=\sum\nolimits_{i=0}^d x_iy_i$ 
 and norm $\lVert \mathbf{x} \rVert=\sqrt{\sum\nolimits_{i=0}^d x_i^2}$.
The tangent space at point $\mathbf{z}$ is $T_\mathbf{z}\mathbb{S}^d=\{\mathbf{v} \in \mathbb{R}^{d+1}  | \langle \mathbf{z}, \mathbf{v} \rangle=0\}$. 
Similar to Lorentz model, we have $\operatorname{Proj}_\mathbf{z}(\mathbf{u}) = \mathbf{u} - \langle \mathbf{z}, \mathbf{u} \rangle\mathbf{z}$
projecting a vector $u \in \mathbb{R}^{d+1}$ into $T_\mathbf{z}\mathbb{S}^d$. 

The exponential map in the sphere model is given as
\begin{align}
    \operatorname{Exp}_\mathbf{z}(\mathbf{v}) = \cos({\lVert \mathbf{v} \rVert})\mathbf{z} + \frac{\sin(\lVert \mathbf{v} \rVert)}{\lVert \mathbf{v} \rVert}\mathbf{v},
    \label{sphere-exp}
\end{align}
and the logarithmic map is
\begin{align}
    \operatorname{Log}_\mathbf{z}(\mathbf{x}) = \frac{\arccos(\langle \mathbf{z}, \mathbf{x} \rangle)}{\sin(\arccos(\langle \mathbf{z}, \mathbf{x} \rangle))}(\mathbf{x} - \langle \mathbf{z}, \mathbf{x} \rangle\mathbf{z})
\end{align}

\paragraph{Derivation of Jacobian}
We first calculate the Jacobian of $\psi$, and it is given as
\begin{align}
    D_\mathbf{z}\psi = \cos({\lVert \mathbf{v} \rVert}) \mathbf{I}.
\end{align}
Similar to that in Lorentz model, the Jacobian of $\phi$ needs the Jacobian of $\lVert \mathbf{v} \rVert$,
\begin{align}
    D_\mathbf{v}\lVert \mathbf{v} \rVert = \frac{d}{d\langle \mathbf{v}, \mathbf{v} \rangle}(\sqrt{\langle \mathbf{v}, \mathbf{v} \rangle})D_\mathbf{v}(\langle \mathbf{v}, \mathbf{v} \rangle)
    = \frac{1}{\lVert \mathbf{v} \rVert}\mathbf{v}^T.
\end{align}
Then, we compute the derivative of the first term of  Eq. \ref{sphere-exp}.
\begin{align}
    D_\mathbf{v}\cos({\lVert \mathbf{v} \rVert})\mathbf{z} = \frac{d}{d\lVert \mathbf{v} \rVert}(\cos({\lVert \mathbf{v} \rVert}))\mathbf{z}(D_\mathbf{v}\lVert \mathbf{v} \rVert)
&=\frac{-\sin(\lVert \mathbf{v} \rVert)}{\lVert \mathbf{v} \rVert}\mathbf{z}\mathbf{v}^T,
\end{align}
and the derivative of the second term is
\begin{align}
    D_\mathbf{v}\frac{\sin(\lVert \mathbf{v} \rVert)}{\lVert \mathbf{v} \rVert}\mathbf{v} &= D_\mathbf{v}(\frac{\sin(\lVert \mathbf{v} \rVert)}{\lVert \mathbf{v} \rVert})\mathbf{v} + \frac{\sin(\lVert \mathbf{v} \rVert)}{\lVert \mathbf{v} \rVert}D_\mathbf{v}\mathbf{v} \\
    &=\frac{\lVert \mathbf{v} \rVert\cos(\lVert \mathbf{v} \rVert) - \sinh(\lVert \mathbf{v} \rVert)}{\lVert \mathbf{v} \rVert^3}\mathbf{v}\mathbf{v}^T + \frac{\sin(\lVert \mathbf{v} \rVert)}{\lVert \mathbf{v} \rVert}\mathbf{I}
\end{align}
Summing up above equations, we finally have
\begin{align}
    D_\mathbf{v}\phi = \frac{\lVert \mathbf{v} \rVert\cos(\lVert \mathbf{v} \rVert_\mathcal{L}) - \sin(\lVert \mathbf{v} \rVert)}{\lVert \mathbf{v} \rVert^3}\mathbf{v}\mathbf{v}^T + \frac{\sin(\lVert \mathbf{v} \rVert)}{\lVert \mathbf{v} \rVert}(\mathbf{I} - \mathbf{z}\mathbf{v}^T).
\end{align}

\section{Riemannian Geometry}
\label{append. riemannian}

\subsection{Some Notations}
Here, we give the formal descriptions of the notions mentioned in the main paper, and please refer to \cite{Petersen16} for systematic elaborations.

\paragraph{Geodesically Complete Manifold.}
A manifold is said to be geodesically complete if  the maximal defining interval of any geodesic is $\mathbb R$. 
For any Riemannian manifold $(\mathcal M, g)$ admitting  a metric structure given by the length of geodesic
\begin{equation}
d(p, q) = \operatorname{inf}\{L(\gamma) | \gamma \text{ is a piecewise smooth curve connecting } p \text{ to } q\},
\end{equation}
 the completeness of $d$ can be described as a metric space is complete if any Cauchy sequence in it converges. 
 For instance, hyperbolic space as well as hyperspherical space is geodesically complete.

\paragraph{Tangent Bundle.}
Given an $n$-dimensional smooth manifold $\mathcal M$, the tangent bundle $T\mathcal M$  is the disjoint union of all the tangent spaces of the manifold $T\mathcal M=\bigsqcup_{\mathbf z \in \mathcal M}T_{\mathbf z}\mathcal M$,
and the tangent bundle with the projection $\pi(\mathbf v)=\mathbf p$ for all $\mathbf v \in T_{\mathbf p}\mathcal M$ is a vector bundle of rank $n$.

\paragraph{Chart.}
A chart of a manifold is a pair $(U, \phi)$ where $U$ is an open set in the manifold and $\phi: U \to \mathbb R^n$ is homeomorphism onto it image, giving a local coordinate of the manifold. In other words, it provides a way of identifying the manifold locally with a Euclidean space.
Given two charts $(U_1, \phi_1)$ and $(U_2, \phi_2)$, if the overlap 
\begin{equation}
\phi_2 \circ \phi_1^{-1}: \phi_1(U_1\cap U_2) \to \phi_2(U_1\cap U_2) \text{ and } \phi_1 \circ \phi_2^{-1}: \phi_2(U_1\cap U_2) \to \phi_1(U_1\cap U_2) ,
\end{equation}
the two charts are said to be compatible.

\paragraph{Curvature and Sectional Curvature.}
The curvature is a notion describing the extent of how a manifold derivatives from being ``flat''. 
In particular, the curvature of a Riemannian manifold $\mathcal M$ should be viewed as a measure $R(X, Y)Z$ of the extent to which the operator 
$(X,Y) \to \nabla_X \nabla_YZ$ is symmetric, where $\nabla$ is a connection on $\mathcal M$ (where $X, Y, Z$ are vector fields, with $Z$ fixed).
Sectional curvature is simpler object of curvature and is defined on two independent vector unit in the tangent space.
When $\nabla$ is the Levi-Civita connection induced by a Riemannian metric on $\mathcal M$, it turns out that the curvature operator $R$ can be recovered from the sectional curvature.

\paragraph{Constant Curvature Space, Hyperbolic Space, Hyperspherical Space.}
A Riemannian manifold is said to be a constant curvature space (CCS) if the sectional curvature is constant scalar everywhere on the manifold.
When the CCS has a negative constant curvature, it is referred to as hyperbolic space, and the CCS is hyperspherical when its constant curvature is positive.

\subsection{$\kappa$-stereographic model and Stereographic Projection}

\paragraph{$\kappa$-stereographic model}

It gives a unified formalism for both positive and negative constant curvatures.
For a positive curvature, it is the hyperspherical model for the hyperspherical space, 
and for a negative curvature, it switches to the Poincar\'{e} ball model.


Specifically, for a curvature $\kappa$ and a dimension $d \geq 2$, 
the $\kappa$-stereographic model $\mathfrak{st}_{\kappa }^{d}$ is defined on the manifold of 
$\{ \mathbf{x} \in \mathbb{R}^d | -\kappa \rVert \mathbf{x}\lVert^2 < 1 \} $,
which is equipped with a Riemannian metric $\mathfrak g_{\mathbf{x}}^{\kappa } =\frac{4}{({1+\kappa \left \| \mathbf{x} \right \|^{2}} )^2  }\mathbf {I} $
for any constant curvature $\kappa$. 
When $\kappa \geq 0$, the  defining domain is  $\mathbb{R}^d$ in which the stereographic projection of the Sphere model of hyperspherical space is endowed.
When $\kappa < 0$, the manifold $\mathfrak{st}_{\kappa }^{d}$  is represented in  an open ball of radius $\frac{1}{\sqrt{-\kappa}}$, and is the stereographic projection of the Lorentz model of hyperbolic space.

The $\kappa$-stereographical model is a gyrovector space in which a non-associative vector operator system is defined.
For $\mathbf{x}, \mathbf{y} \in \mathbb{G}^n$, $a \in \mathbb{R}$, the $\kappa$-addition (a.k.a. M\"{o}bius addition) is given as
\begin{align}
    \mathbf{x} \oplus_\kappa \mathbf{y} = \frac{(1-2\kappa\mathbf{x}^T\mathbf{y}-\kappa\lVert \mathbf{y} \rVert^2)\mathbf{x}+(1+\kappa\lVert \mathbf{x} \rVert^2)\mathbf{y}}{1-2\kappa\mathbf{x}^T\mathbf{y}+\kappa^2\lVert \mathbf{x} \rVert^2\lVert \mathbf{y} \rVert^2}
\end{align}
The distance function given by $\kappa$-addition is thus formulated as 
\begin{align}
    d_\kappa(\mathbf{x}, \mathbf{y})=2\tan_\kappa^{-1}(\lVert (-\mathbf{x}) \oplus_\kappa \mathbf{y} \rVert)
\end{align}
The $\kappa$-scaling for any real scalar $c$ is defined as 
\begin{align}
    c \otimes_\kappa \mathbf{x} = \tan_\kappa(c \cdot \tan^{-1}_\kappa(\lVert \mathbf{x} \rVert))\frac{\mathbf{x}}{\lVert \mathbf{x} \rVert}
\end{align}
The unit-speed geodesic from $\mathbf{x}$ to $\mathbf{y}$ is
\begin{align}
    \gamma_{\mathbf{x} \rightarrow \mathbf{y}}(t)=\mathbf{x} \oplus_\kappa(t\otimes_\kappa((-\mathbf{x}) \oplus_\kappa\mathbf{y}))
\end{align}
With the unit-speed geodesic, the exponential map as well as its inverse (i.e., the logarithmic map) has the closed-form expression as follows,
\begin{align}
    \operatorname{Exp}_\mathbf{x}^\kappa(\mathbf{v})&=\mathbf{x} \oplus_\kappa(\tan_\kappa(\lvert \kappa\rvert^{\frac{1}{2}}\frac{\lambda^\kappa_\mathbf{x}\lVert \mathbf{v} \rVert}{2})\frac{\mathbf{v}}{\lVert \mathbf{v} \rVert}) \\
    \operatorname{Log}_\mathbf{x}^\kappa(\mathbf{y})&=\frac{2\lvert \kappa\rvert^{-\frac{1}{2}}}{\lambda^\kappa_\mathbf{x}}\tan_\kappa^{-1}(\lVert (-\mathbf{x}) \oplus_\kappa \mathbf{y} \rVert)\frac{(-\mathbf{x}) \oplus_\kappa \mathbf{y}}{\lVert (-\mathbf{x}) \oplus_\kappa \mathbf{y} \rVert},
\end{align}
where the curvature-aware trigonometric function is utilized, e.g.,
        \begin{equation}
            \tan_\kappa(x)=
            \begin{cases}
                \frac{1}{\sqrt{\kappa}}\tan(x) &\kappa > 0, \\
                x &\kappa = 0, \\
                \frac{1}{\sqrt{-\kappa}}\tanh(x) &\kappa < 0. \\
            \end{cases}
        \end{equation}

\paragraph{Stereographic Projection}

The stereographic projection is a diffeomorphism  connecting the different model spaces of Riemannian manifold.
In particular, it is defined as a map $\pi: \mathbb{L}^d_\kappa/\mathbb{S}^d_\kappa \rightarrow \mathfrak{st}_{\kappa }^{d}$ taking the form of
\begin{align}
\pi: \mathbb{L}^d_\kappa/\mathbb{S}^d_\kappa \rightarrow \mathfrak{st}_{\kappa }^{d},  \quad \mathbf{x}=\frac{1}{1+\sqrt{|\kappa|} \mathbf{x}_{d+1}^{\prime}} \mathbf{x}_{1: d}^{\prime},
\end{align}
where \( \mathbf{x}' \) is a point on the Lorentz model $\mathbb{L}^d_\kappa$ or Sphere model $\mathbb{L}^d_\kappa$, 
and  \( \mathbf{x} \), the image of the projection,  is the corresponding point in the gyrovector ball of $\kappa-$stereographic model.
The inverse  projection is given as follows, 
\begin{align}
\pi^{-1}:   \mathfrak{st}_{\kappa }^{d}\rightarrow \mathbb{L}^d_\kappa/\mathbb{S}^d_\kappa , \quad \mathbf{x}^{\prime}=
\left ( \lambda _{\mathbf{x}}^{\kappa}\mathbf{x}, \frac{1}{\sqrt{|\kappa| } }  \left ( \lambda_{\mathbf{x}}^{\kappa}-1  \right )  \right ), 
\end{align}
where \( \lambda _{\mathbf{x}}^{\kappa}=\frac{2}{1+\kappa \left \| \mathbf{x} \right \|^{2}  }  \) is known as the conformal factor.


\subsection{Cartesian Product and Product Manifold}\label{product}

The concept of product manifolds allows for creating a new manifold from a finite collection of existing ones. 
Given a set of smooth manifolds $\mathcal{M}_1, \mathcal{M}_2, \ldots, \mathcal{M}_k$, 
the product manifold $\mathbb{P}$ is given as the Cartesian product of these manifolds: 
\begin{equation}
\mathbb{P} = \mathcal{M}_1 \times \mathcal{M}_2 \times \ldots \times \mathcal{M}_k,
\end{equation} 
where $\otimes$ denotes the Cartesian product.
Specifically, with the Cartesian  product construction, 
a point $\mathbf x \in \mathbb{P}$ are represented by a concatenation of  $\mathbf{x}=[\mathbf{x}_{1} ,\dots,\mathbf{x}_{k}]$, 
where $\mathbf{x}_{i}\in \mathcal{M}_{i}$. 
A tangent vector $\mathbf v \in T_{\mathbf x}\mathbb{P}$ at a point $\mathbf{x}$ can be given as $\mathbf{v}=[\mathbf{v}_{1} ,\dots,\mathbf{v}_{k}]$, where $\mathbf{v}_{i}\in T_{\mathbf{x}_{i}}\mathcal{M}_{i}$.
If each manifold $\mathcal{M}_i$ is equipped with a metric tensor $\mathbf{g}_i$,
 the product metric $\mathbf{g}$ decomposes into the direct sum of the individual metrics 
$ \mathbf{g}=  \oplus_{i=1}^k \mathbf{g}^{i}$, 
which can be expressed as $\operatorname{Diag}(\mathbf{g}^1,...,\mathbf{g}^k)$. 
For $\mathbf{x}$ and $\mathbf{y} \in \mathbf{P}$, the distance between them is defined as $d_\mathbb{P}(\mathbf{x}, \mathbf{y})=\sum_{i=1}^k d_{\mathcal{M}_i}(\mathbf{x}_i, \mathbf{y}_i)$.
Accordingly, the exponential map is given as 
\begin{equation}
\operatorname{Exp}_{\mathbf x}([\mathbf{v}_{1} ,\dots,\mathbf{v}_{k}])
=\left[ \operatorname{Exp}_{\mathbf{x}_{1}}(\mathbf{v}_{1}) , \operatorname{Exp}_{\mathbf{x}_{2}}(\mathbf{v}_{2}), \dots, \operatorname{Exp}_{\mathbf{x}_{k}}(\mathbf{v}_{k}) \right].
\end{equation}

\section{Experimental Setups}
\label{append. exp}

\begin{table}[h]
    \centering
    \caption{Dataset statitics.}
    \begin{tabular}{l c c c c }
    \hline
        \textbf{} &  \textbf{Computers} & \textbf{Photo} & \textbf{CS} & \textbf{Physics}  \\ \hline
        \textbf{\#Nodes}  & 13752 & 7650  &18,333   & 34,493\\ 
        \textbf{\#Features}  & 767 & 745 & 6,805  & 8,415 \\ 
        \textbf{\#Edges} & 245861 & 119081  & 163,788  & 495,924\\ 
        \textbf{\#Classes } & 10  & 8   & 15   & 5\\ \hline
    \end{tabular}
    \label{table:Dataset statitics}
\end{table}

\subsection{Dataset}
\label{append. data}
We use four common benchmark datasets to evaluate our model and Table \ref{table:Dataset statitics} show the details of the datasets. There are two co-purchase graphs including  Amazon-Photo and Amazon-Computers \cite{shchur2018pitfalls} and two co-author network including CS and Physics \cite{shchur2018pitfalls}.

\subsection{Baselines}

\begin{table}[!ht]
    \centering
    \caption{The categories of the baselines}
    \begin{tabular}{ |r |c| c| }
    \hline
        \textbf{} & \textbf{ANN-based Models} & \textbf{SNN-based Models}  \\ 
        \hline
        \textbf{Euclidean Space} & 
        \makecell[c]{GCN \cite{kipf2016semi}, SAGE \cite{hamilton2017inductive}, \\GAT \cite{velickovic2018graph}, SGC \cite{wu2019simplifying} }
        &  \makecell[c]{SpikeNet \cite{li2023scaling}, SpikeGraphormer \cite{sun2024spikegraphormer}, \\ SpikeGCN \cite{zhu2022spiking}, SpikeGCL \cite{li2023graph} \\}  \\ 
         \hline
        \textbf{Riemannian Space} & \makecell[c]{ $Q-$GCN \cite{xiong2022pseudo}, $\kappa-$GCN \cite{bachmann2020constanta}, \\ HyboNet \cite{chen2021fully}, HGCN \cite{chami2019hyperbolic} }& \\ 
            \hline
    \end{tabular}
    \label{table:baselines}
\end{table}

As shown in Table \ref{table:baselines}, we divide the baselines into three categories: ANN-based Euclidean GNNs, ANN-based Riemannian GNNs and SNN-based Euclidean GNNs.  Note that, none of the existing work studies the SNN-based GNN in Riemannian space, to the best of our knowledge.
\paragraph{ANN-based Euclidean GNNs}
\begin{itemize}
\item \textbf{GCN} \cite{kipf2016semi}: It defines graph convolution on the spectral domain.
\item \textbf{SAGE} \cite{hamilton2017inductive}: It gives the aggregate-and-combine formulation for the message passing over the graph.
\item \textbf{GAT} \cite{velickovic2018graph}: It introduces the attention mechanism for the learning on graphs.
\item \textbf{SGC} \cite{wu2019simplifying}: It reformulates GCN \cite{kipf2016semi} with feature propagation and linear layer, acting as a low-pass filter.
\end{itemize}
\paragraph{ANN-based Riemannian GNNs}
\begin{itemize}
\item \textbf{HGCN} \cite{chami2019hyperbolic} : It generalizes GAT \cite{velickovic2018graph} in the Lorentz model of hyperbolic space in which the graph convolution is conducted in the tangent space.
\item \textbf{$\kappa-$GCN} \cite{bachmann2020constanta}: It generalizes GCN \cite{kipf2016semi} to the $\kappa-$stereographical model of constant curvature spaces where several gyrovector operators are given in the unified formalism.
\item \textbf{HyboNet} \cite{chen2021fully}: It introduces a parameterized Lorentz transformation for hyperbolic graph modeling without the tangent space.
\item \textbf{$\mathcal{Q}-$GCN} \cite{xiong2022pseudo}: It studies the graph convolution network in the Pseudo-Riemannian manifold.
\end{itemize}
\paragraph{SNN-based Euclidean GNNs. }
\begin{itemize}
\item \textbf{SpikeGCN} \cite{zhu2022spiking}: It integrates SNN and graph convolution network in which SNN acts as an activation function.
\item \textbf{SpikeGraphormer} \cite{sun2024spikegraphormer} (termed as SpikeGT for short in our main paper): It generalizes a kind of graph transformer with spiking neurons, accompanied by an ANN for improving the performance.
\item \textbf{SpikeGCL} \cite{li2023graph}: It introduces a method to perform graph contrastive learning with the spiking GNN.
\item \textbf{SpikeNet} \cite{li2023scaling}: It is original designed for dynamic graph modeling, and we utilize its version for static graph following \cite{li2023graph}.
\end{itemize}
Note that, existing SNN-based GNNs work with Euclidean space, and leverage the BPTT training with surrogate gradient, suffering from high latency.

\subsection{Theoretical Energy Consumption}
Following the previous works \cite{li2023graph,zhu2022spiking,zhou2022spikformer}, we calculate the theoretical energy consumption for each model, instead of measuring actual electricity usage, for fair comparison.
\begin{itemize}
\item  
For the SNN-based models, the energy consumption involves encoding energy $E_{\text{encoding}}$ and spiking process energy $E_{\text{spiking}}$. 
The former is calculated by the number of multiply-and-accumulate (MAC) operations, and the latter is given by the number of SOP operations. 
The energy consumption is thus defined as follows,
\begin{align}
E=E_{\text{encoding}}+E_{\text{spiking}}=E_{\mathrm{MAC}}\sum_{t=1}^{T}Nd S_t+E_{\mathrm{SOP}}\sum_{t=1}^{T}\sum_{l=1}^{L}S_{t}^{l},
\end{align}
where the scaling constant $E_{\mathrm{MAC}}$ and $E_{\mathrm{SOP}}$ are set to $4.6pJ$ and $3.7pJ$, respectively.
$N$ is the number of nodes in the graph, $d$ is the dimension of node features, $L$ is the number of layers in the neural model.
$T$ is the time steps of the spikes, and  $S_{t}^{l}$ denotes the output spikes at time step $t$ and layer $l$. 
\item 
For the ANN-based models, the energy consumption is given by embedding generation step and aggregation step, 
and both of them  are calculated by MAC operations. 
In the embedding generation step, the feature transformation with a weight matrix of  $\mathbb{R}^{d_{in}\times d_{out}}$ executes $Nd_{in}d_{out}$ multiplication and $Nd_{in}d_{out}$  addition operations. 
In the aggregation step, $|\mathcal E|d_{in}$ is the number of multiplication and $|\mathcal E|d_{out}$ is the number of addition operations. 
Supposing $|\mathcal E|d_{in}=|\mathcal E|d_{out}$, the energy consumption is given as follows,
\begin{align}
    E=E_{\mathrm{MAC}}(Nd_{in}d_{out} + |\mathcal E|d_{out}).
\end{align}
where the scaling constant $E_{\mathrm{MAC}}$ is set to $4.6pJ$, $|\mathcal E|$ is the number of edge, $d_{in}$ and $d_{out}$ are the input and output dimensions. 
\end{itemize}


\subsection{Implementation Notes}

\paragraph{Model Instantiation.} 
The proposed \texttt{MSG} is instantiated  in the Lorentz model $\mathbb H$ of hyperbolic space  or  Sphere model $\mathbb S$ of hyperspherical  space as well as the products over $\mathbb H$ and $\mathbb S$. 
Note that, \texttt{MSG} can be equivalently instantiated on the $\kappa-$stereographic model (i.e., Poincar\'{e} model of hyperbolic space  or  Hyperspherical model  of hyperspherical  space), given the closed form  Riemannian metric and exponential map. The equivalence can be achieved by scaling the stereographical projection.
We leverage the  Lorentz model $\mathbb H$ by default.

\paragraph{Hyperparameters.}
The dimension of the manifold is set as $32$. When we instantiate \texttt{MSG} on the product manifold, the sum of factor manifold's dimensions is defined as $32$.
The manifold spiking neuron is based on the IF model \cite{burkitt2006review} by default, and it is ready to switch to the LIF model \cite{burkitt2006review}.
The time latency $T$ for neurons is set to $5$ or $15$. 
The step size $\epsilon$ in Eq. \ref{eq. z} is set  to $0.1$.
The hyperparameters are tuned with grid search, in which the learning rate is $\{0.01, 0.003\}$ for node classification and $\{0.003, 0.001\}$ for link prediction, and the dropout rate is in $\{0.1, 0.3, 0.5\}$. 

\paragraph{Hardware.} Experiments are conducted on the NVIDIA GeForce RTX 4090 GPU 24GB memory, and AMD EPYC 9654 CPU
315 with 96-Core Processor.

\section{Additional Results}
\label{append.add}

In this section, we show the additional results on backward gradient, comparison between IF and LIF model, link prediction and visualization.


\begin{figure}[h]
    \centering
    \subfigure[The norm of backward gradient of $L$ with respect to $\mathbf{z}$ in each spiking layers.]{
        \includegraphics[width=1\textwidth]{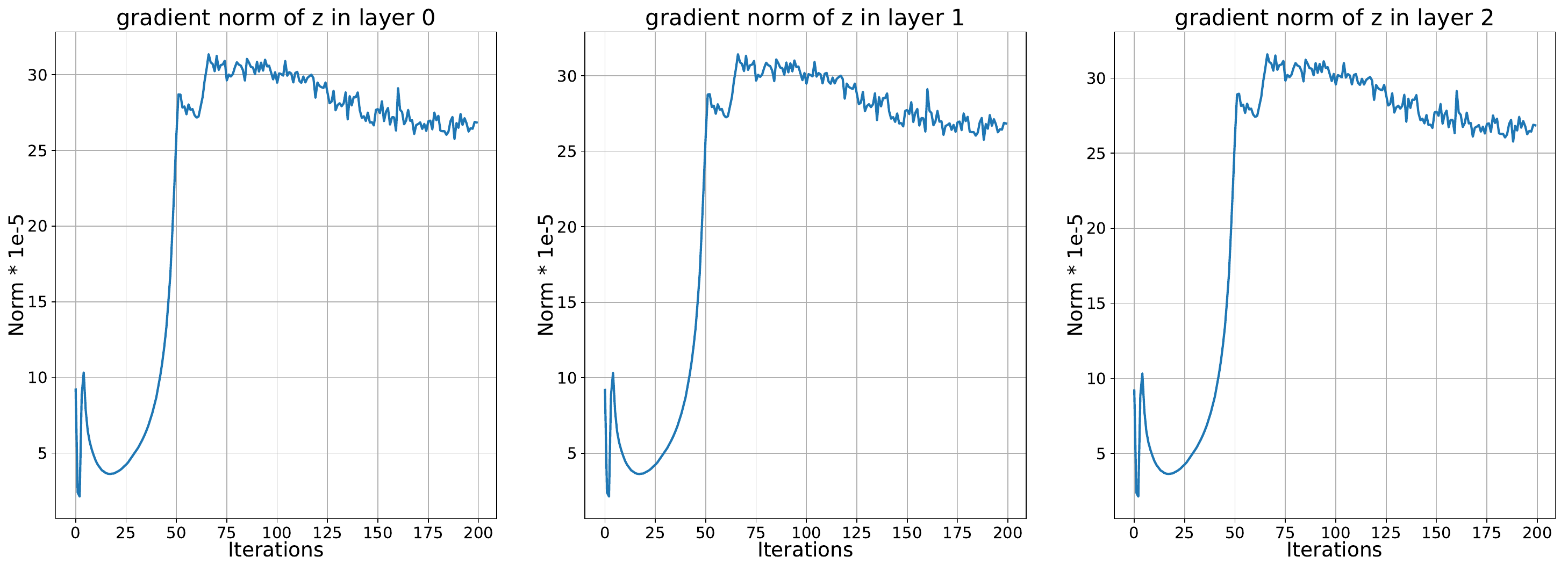}
    }
    \subfigure[The norm of backward gradient of $L$ with respect to $\mathbf{v}$ in each spiking layers.]{
        \includegraphics[width=1\textwidth]{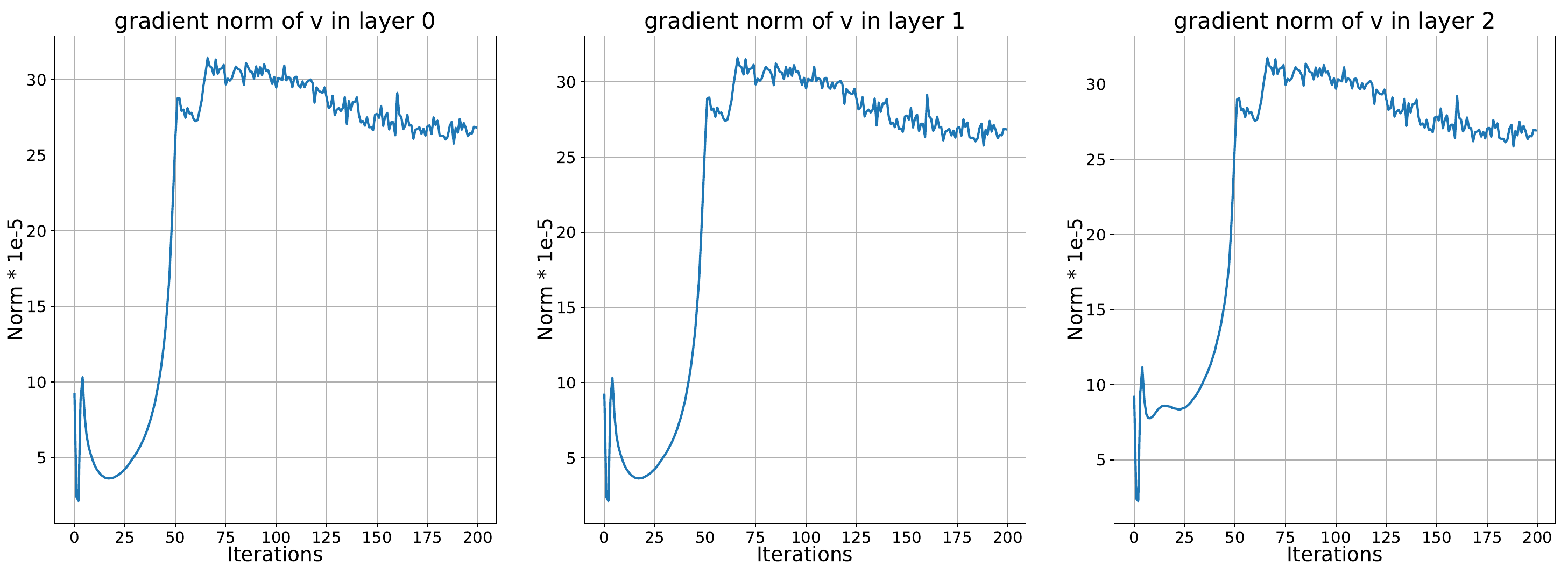}
    }
    \subfigure[The loss in model training.]{
        \includegraphics[width=0.333\textwidth]{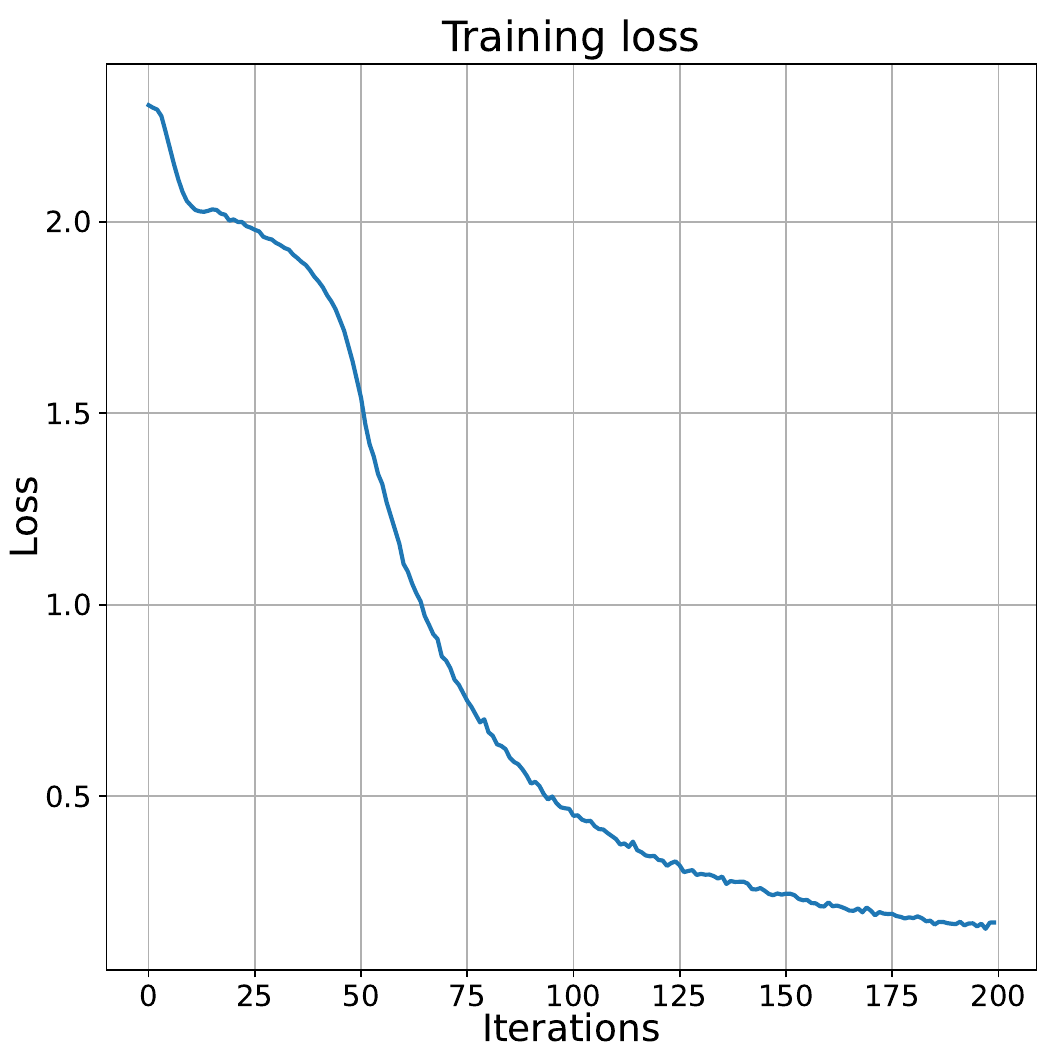}
    }
    \caption{Visualizations of the training process for node classification on Computer dataset.}
    \label{fig:fig_micPerMon}
    \vspace{-0.1in}
\end{figure}

\paragraph{Backward Gradient.}
Previous studies compute backward gradients though the Differentiation via Spikes (\emph{DvS}).
Distinguishing from the previous studies, we compute backward gradients though the Differentiation via Manifold (\emph{DvM}).
In order to examine the backward gradients, we visualize the training process for node classification on Computer dataset.
Concretely, we plot the norm of backward gradients in each iteration in Figs. \ref{fig:fig_micPerMon} (a) and (b) together with the value of loss function in Figs. \ref{fig:fig_micPerMon} (c). 
As shown in Fig. \ref{fig:fig_micPerMon}, the proposed algorithm with \emph{DvM} converges well, 
and \emph{the backward gradients  do not suffer from gradient vanishing or gradient explosion.}

\paragraph{Comparison between IF and LIF model.}
In the main paper, the proposed \texttt{MSG} is built with the IF model, and it is applicable to LIF model as well.
We compare the performance between IF and LIF model in different algorithms (\emph{DvM} and \emph{DvS}) and in different manifolds (hyperbolic $\mathbb H$, hyperspherical $\mathbb S$, Euclidean $\mathbb E$ and the product spaces among $\mathbb H$ and $\mathbb S$) for a comprehensive evaluation.
The results of node classification on Computer, Photo, CS and Physics datasets are summarized in Table  \ref{tab. node} and Table \ref{tab. msg_node}.
Note that, \emph{IF model and LIF model achieves competitive performance in every case. 
We opt for IF model in the model instantiation for simplicity.}

\begin{table}[h]
\centering
\caption{Comparison between IF and LIF model in Node Classification, qualified by classification accuracy (\%). The proposed model is trained by Differentiation via Spikes (i.e., BPTT with the surrogate gradient).}
\label{tab. node}
\begin{tabular}{l c|c c c c}
      \hline
        &  & \multicolumn{1}{c}{\textbf{Computers}}  & \multicolumn{1}{c}{\textbf{Photo}}  & \multicolumn{1}{c}{\textbf{CS}} & \multicolumn{1}{c}{\textbf{Physics}} \\
      \hline
      \multirow{3}{*}{\rotatebox{90}{IF}} 
& $\mathbb{H}^{32}$  & 89.65$\pm$0.18              & 93.46$\pm$0.12              & 91.73$\pm$0.34 & 95.16$\pm$0.17 \\
& $\mathbb{S}^{32}$  & 89.37$\pm$0.26              & 93.39$\pm$0.21              & 91.59$\pm$0.24 & 95.15$\pm$0.10 \\
& $\mathbb{E}^{32}$  & 88.36$\pm$0.95              & 92.75$\pm$0.40              & 92.53$\pm$0.06 & 96.00$\pm$0.03 \\
\hline
\multirow{3}{*}{\rotatebox{90}{LIF}} 
& $\mathbb{H}^{32}$ & 89.58$\pm$0.34              & 92.81$\pm$0.21              & 92.44$\pm$0.13 & 95.63$\pm$0.02 \\
& $\mathbb{S}^{32}$ & 89.32$\pm$0.19              & 92.82$\pm$0.15              & 92.11$\pm$0.16 & 95.54$\pm$0.04 \\
& $\mathbb{E}^{32}$ & 89.13$\pm$0.27              & 92.93$\pm$0.23              & 92.56$\pm$0.15 & 95.97$\pm$0.05 \\
\hline
\end{tabular}
\end{table}

\begin{table}[h]
\centering
\caption{Comparison between IF and LIF model in Node Classification, qualified by classification accuracy (\%). The proposed model is trained by Differentiation via Manifold.}
\label{tab. msg_node}
\begin{tabular}{l c|c c c c}
      \hline
        &  & \multicolumn{1}{c}{\textbf{Computers}}  & \multicolumn{1}{c}{\textbf{Photo}}  & \multicolumn{1}{c}{\textbf{CS}} & \multicolumn{1}{c}{\textbf{Physics}} \\
      \hline
\multirow{3}{*}{\rotatebox{90}{IF}} 
&  $\mathbb{H}^{32}$ & 89.27$\pm$0.19              & 93.11$\pm$0.11              & 92.65$\pm$0.04    & 95.93$\pm$0.07 \\
& $\mathbb{S}^{32}$  & 87.84$\pm$0.77              & 92.03$\pm$0.79              & 92.72$\pm$0.06    & 95.85$\pm$0.02 \\
& $\mathbb{E}^{32}$  & 88.94$\pm$0.24              & 92.93$\pm$0.21              & 92.82$\pm$0.04    & 95.81$\pm$0.04 \\
\hline
\multirow{3}{*}{\rotatebox{90}{LIF}} 
& $\mathbb{H}^{32}$ & 88.71$\pm$0.13              & 92.74$\pm$0.07            & 92.43$\pm$0.11     & 95.64$\pm$0.06            \\
& $\mathbb{S}^{32}$ & 88.43$\pm$0.10              & 92.45$\pm$0.13             & 92.53$\pm$0.06    & 95.84$\pm$0.03            \\
& $\mathbb{E}^{32}$ & 86.34$\pm$0.19              & 92.42$\pm$0.09            & 92.66$\pm$0.09     & 96.02$\pm$0.03       \\
\hline
\end{tabular}
\end{table}

\paragraph{Link Prediction.}
The performance of link prediction in terms of AUC is provided in the main paper.
We show the results on Computer, Photo, CS and Physics datasets in terms of AP (\%) in Table \ref{table. lp_ap}. 
In particular, we feed the spiking representation of SpikingGCN and SpikeGCL into the Fermi-Dirac decoder same as the proposed \texttt{MSG},
while SpikeNet and SpikeGT are designed for node classification specially.
As shown in Table \ref{table. lp_ap}, \emph{the proposed spiking  \texttt{MSG} consistently achieves the best results among the spiking GNNs on all the four datasets, and even achieves the competitive performances with the strong Riemannian baselines.}

\begin{table}[h]
    \caption{Link Prediction in terms of AP (\%) on Computers, Photo, CS and Physics datasets. The best results are \textbf{boldfaced}, and the runner-ups are \underline{underlined}. The standard derivations are given in the subscripts. OOM denotes Out-Of-Memory.}
        \centering
    \begin{tabular}{cl|c c  c c}
    \hline 
        &  & \multicolumn{1}{c}{\textbf{Computers}}  & \multicolumn{1}{c}{\textbf{Photo}}  & \multicolumn{1}{c}{\textbf{CS}} & \multicolumn{1}{c}{\textbf{Physics}} \\
    \hline
     \multirow{4}{*}{\rotatebox{90}{\scriptsize  ANN-\textbf{E}}} 
       & GCN \cite{kipf2016semi} 
       & 92.10{\scriptsize $\pm$0.50} & 86.43{\scriptsize $\pm$0.40} & 93.38{\scriptsize $\pm$0.92} & 92.83{\scriptsize $\pm$0.47}  \\ 
       & GAT \cite{velickovic2018graph} 
       & 91.61{\scriptsize $\pm$0.65} & 87.04{\scriptsize $\pm$0.05} & 94.34{\scriptsize $\pm$0.60} & 93.44{\scriptsize $\pm$0.45} \\ 
       & SGC \cite{wu2019simplifying} 
       & 90.78{\scriptsize $\pm$0.60} & 90.05{\scriptsize $\pm$0.70} & \underline{95.34{\scriptsize $\pm$0.58}} & \underline{95.37{\scriptsize $\pm$0.81}} \\ 
       & SAGE \cite{hamilton2017inductive} 
       & 90.51{\scriptsize $\pm$0.42} & 88.40{\scriptsize $\pm$0.40} & 94.86{\scriptsize $\pm$0.21} & 95.15{\scriptsize $\pm$0.51}  \\ 
    \hline
       \multirow{4}{*}{\rotatebox{90}{\scriptsize  ANN-\textbf{R}}} 
       & HGCN \cite{chami2019hyperbolic} 
       & \textbf{96.46{\scriptsize $\pm$0.74}} & 93.86{\scriptsize $\pm$0.30} & 91.90{\scriptsize $\pm$0.35} & 92.01{\scriptsize $\pm$0.57}  \\ 
       & $\kappa$-GCN \cite{bachmann2020constanta} 
       & 94.80{\scriptsize $\pm$0.60} & 93.50{\scriptsize $\pm$0.09} & 94.97{\scriptsize $\pm$0.07} & 94.16{\scriptsize $\pm$0.48} \\ 
       & $\mathcal{Q}$-GCN \cite{xiong2022pseudo} 
       & \underline{96.28{\scriptsize $\pm$0.03}} & \underline{96.65{\scriptsize $\pm$0.10}} & 92.24{\scriptsize $\pm$0.75} & OOM \\ 
       & HyboNet \cite{chen2021fully} 
       & 95.78{\scriptsize $\pm$0.07} & \textbf{96.79{\scriptsize $\pm$0.04}} & \textbf{96.21{\scriptsize $\pm$0.33}} & \textbf{98.12{\scriptsize $\pm$0.97}} \\ 
    \hline
       \multirow{4}{*}{\rotatebox{90}{\scriptsize  SNN-\textbf{E}}} 
       & SpikeNet \cite{li2023scaling} 
       & - & - & - & - \\ 
       & SpikingGCN \cite{zhu2022spiking} 
       & 91.17{\scriptsize $\pm$1.64} & 93.16{\scriptsize $\pm$0.04} & 94.79{\scriptsize $\pm$1.23} & 92.19{\scriptsize $\pm$0.90}  \\ 
       & SpikeGCL \cite{li2023graph} 
       & 92.54{\scriptsize $\pm$0.03} & 95.16{\scriptsize $\pm$0.12} & 95.06{\scriptsize $\pm$0.19} & 91.82{\scriptsize $\pm$0.25}  \\ 
       & SpikeGT \cite{sun2024spikegraphormer} 
       & - & - & - & -\\ 
    \hline
       & \texttt{MSG} (Ours) 
       & 94.45{\scriptsize $\pm$0.78} & 96.46{\scriptsize $\pm$0.19} & 95.12{\scriptsize $\pm$0.12}  & 92.53{\scriptsize $\pm$0.19}  \\ 
       \hline
    \end{tabular}
    \label{table. lp_ap}
\end{table}

\paragraph{Visualization.}
Here, we visualize the forward pass of the proposed \texttt{MSG}  and empirically demonstrate the connection between \texttt{MSG}  and manifold ordinary differential equation (ODE).

\begin{figure}[h]
    \centering
    \subfigure[$1$-th nodes.]{
        \includegraphics[width=0.6\textwidth]{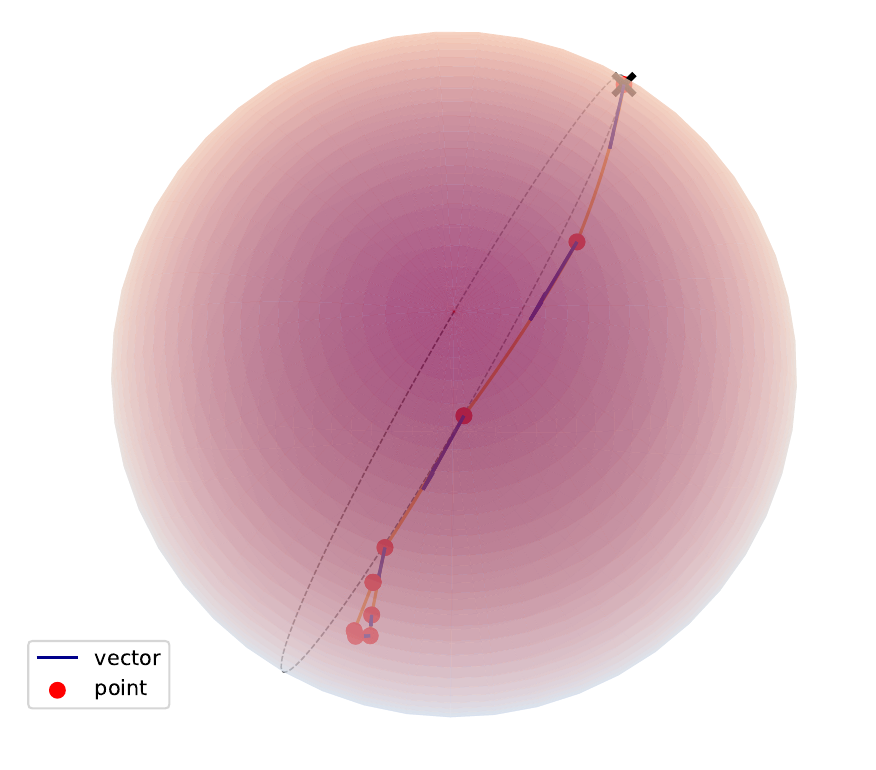}
        \label{fig:manifold_1}
    }
    \subfigure[$17$-th nodes.]{
        \includegraphics[width=0.6\textwidth]{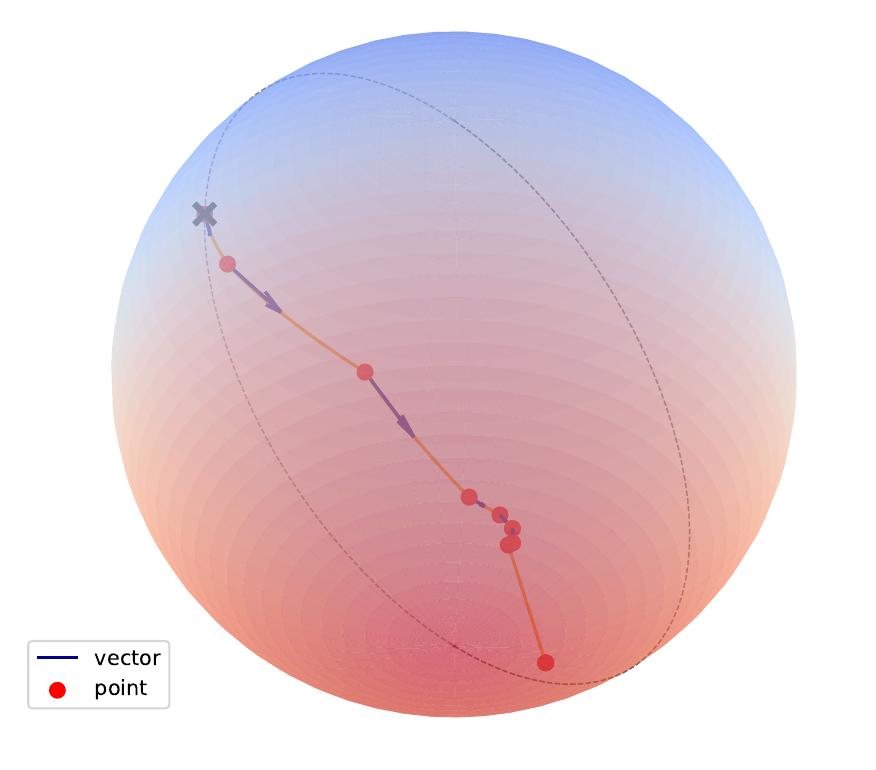}
        \label{fig:manifold_17}
    }
    
    \caption{Visualizations of node representations on Zachary karateClub datasets \cite{zachary1977information}.}
    \label{fig: karateclub}
\end{figure}

We choose a toy example of KarateClub dataset. The proposed \texttt{MSG} are instantiated on the 2D manifold for the ease of visualization. 
Specifically, we plot  the node representation of each spiking layer in Fig. \ref{fig:manifold_1} and Fig. \ref{fig:manifold_17} in which the curve connecting the outputs of successive layer is marked in red, 
and blue arrow is the direction of the geodesic.
It is shown that a spiking layer forwards the node along the geodesic on the manifold. 
In other words, \emph{each layer, constructing a chart given by the exponential map, is a solver of the ODE describing the geodesic.}



\newpage
\section*{NeurIPS Paper Checklist}

\begin{enumerate}

\item {\bf Claims}
    \item[] Question: Do the main claims made in the abstract and introduction accurately reflect the paper's contributions and scope?
    \item[] Answer: \answerYes{} 
    \item[] Justification: We propose a novel Manifold-valued Spiking GNN (\texttt{MSG}), and design a new training algorithm with \textit{Differentiation via Manifold} with theoretical gaurantee. Extensive experiments show the effectiveness of the proposed approach.
    \item[] Guidelines:
    \begin{itemize}
        \item The answer NA means that the abstract and introduction do not include the claims made in the paper.
        \item The abstract and/or introduction should clearly state the claims made, including the contributions made in the paper and important assumptions and limitations. A No or NA answer to this question will not be perceived well by the reviewers. 
        \item The claims made should match theoretical and experimental results, and reflect how much the results can be expected to generalize to other settings. 
        \item It is fine to include aspirational goals as motivation as long as it is clear that these goals are not attained by the paper. 
    \end{itemize}

\item {\bf Limitations}
    \item[] Question: Does the paper discuss the limitations of the work performed by the authors?
    \item[] Answer: \answerYes{} 
    \item[] Justification: We discuss the limitations in Sec. \ref{impact}. The proposed model is applicable to any geodesically complete manifold (e.g., hyperbolic space, hyperspherical space and their products), and its generalization to more generic manifold leaves as the future work.
    \item[] Guidelines:
    \begin{itemize}
        \item The answer NA means that the paper has no limitation while the answer No means that the paper has limitations, but those are not discussed in the paper. 
        \item The authors are encouraged to create a separate "Limitations" section in their paper.
        \item The paper should point out any strong assumptions and how robust the results are to violations of these assumptions (e.g., independence assumptions, noiseless settings, model well-specification, asymptotic approximations only holding locally). The authors should reflect on how these assumptions might be violated in practice and what the implications would be.
        \item The authors should reflect on the scope of the claims made, e.g., if the approach was only tested on a few datasets or with a few runs. In general, empirical results often depend on implicit assumptions, which should be articulated.
        \item The authors should reflect on the factors that influence the performance of the approach. For example, a facial recognition algorithm may perform poorly when image resolution is low or images are taken in low lighting. Or a speech-to-text system might not be used reliably to provide closed captions for online lectures because it fails to handle technical jargon.
        \item The authors should discuss the computational efficiency of the proposed algorithms and how they scale with dataset size.
        \item If applicable, the authors should discuss possible limitations of their approach to address problems of privacy and fairness.
        \item While the authors might fear that complete honesty about limitations might be used by reviewers as grounds for rejection, a worse outcome might be that reviewers discover limitations that aren't acknowledged in the paper. The authors should use their best judgment and recognize that individual actions in favor of transparency play an important role in developing norms that preserve the integrity of the community. Reviewers will be specifically instructed to not penalize honesty concerning limitations.
    \end{itemize}

\item {\bf Theory Assumptions and Proofs}
    \item[] Question: For each theoretical result, does the paper provide the full set of assumptions and a complete (and correct) proof?
    \item[] Answer: \answerYes{} 
    \item[] Justification: We provide the complete proof in Appendix \ref{proof}.
    \item[] Guidelines:
    \begin{itemize}
        \item The answer NA means that the paper does not include theoretical results. 
        \item All the theorems, formulas, and proofs in the paper should be numbered and cross-referenced.
        \item All assumptions should be clearly stated or referenced in the statement of any theorems.
        \item The proofs can either appear in the main paper or the supplemental material, but if they appear in the supplemental material, the authors are encouraged to provide a short proof sketch to provide intuition. 
        \item Inversely, any informal proof provided in the core of the paper should be complemented by formal proofs provided in appendix or supplemental material.
        \item Theorems and Lemmas that the proof relies upon should be properly referenced. 
    \end{itemize}

    \item {\bf Experimental Result Reproducibility}
    \item[] Question: Does the paper fully disclose all the information needed to reproduce the main experimental results of the paper to the extent that it affects the main claims and/or conclusions of the paper (regardless of whether the code and data are provided or not)?
    \item[] Answer: \answerYes{} 
    \item[] Justification: Experimental details are given in Sec. \ref{exp_setup}, and further introduced in Appendix \ref{append. exp}. Also, we give pseudocodes in Algorithm \ref{alg. msg}.
    \item[] Guidelines:
    \begin{itemize}
        \item The answer NA means that the paper does not include experiments.
        \item If the paper includes experiments, a No answer to this question will not be perceived well by the reviewers: Making the paper reproducible is important, regardless of whether the code and data are provided or not.
        \item If the contribution is a dataset and/or model, the authors should describe the steps taken to make their results reproducible or verifiable. 
        \item Depending on the contribution, reproducibility can be accomplished in various ways. For example, if the contribution is a novel architecture, describing the architecture fully might suffice, or if the contribution is a specific model and empirical evaluation, it may be necessary to either make it possible for others to replicate the model with the same dataset, or provide access to the model. In general. releasing code and data is often one good way to accomplish this, but reproducibility can also be provided via detailed instructions for how to replicate the results, access to a hosted model (e.g., in the case of a large language model), releasing of a model checkpoint, or other means that are appropriate to the research performed.
        \item While NeurIPS does not require releasing code, the conference does require all submissions to provide some reasonable avenue for reproducibility, which may depend on the nature of the contribution. For example
        \begin{enumerate}
            \item If the contribution is primarily a new algorithm, the paper should make it clear how to reproduce that algorithm.
            \item If the contribution is primarily a new model architecture, the paper should describe the architecture clearly and fully.
            \item If the contribution is a new model (e.g., a large language model), then there should either be a way to access this model for reproducing the results or a way to reproduce the model (e.g., with an open-source dataset or instructions for how to construct the dataset).
            \item We recognize that reproducibility may be tricky in some cases, in which case authors are welcome to describe the particular way they provide for reproducibility. In the case of closed-source models, it may be that access to the model is limited in some way (e.g., to registered users), but it should be possible for other researchers to have some path to reproducing or verifying the results.
        \end{enumerate}
    \end{itemize}

\item {\bf Open access to data and code}
    \item[] Question: Does the paper provide open access to the data and code, with sufficient instructions to faithfully reproduce the main experimental results, as described in supplemental material?
    \item[] Answer: \answerYes{} 
    \item[] Justification: The datasets are publicly available. We properly cite and introduce the datasets in Sec. \ref{exp_setup} and Appendix \ref{append. data}. Our source code is at the anonymous link \url{https://anonymous.4open.science/r/MSG-16E9}.
    \item[] Guidelines:
    \begin{itemize}
        \item The answer NA means that paper does not include experiments requiring code.
        \item Please see the NeurIPS code and data submission guidelines (\url{https://nips.cc/public/guides/CodeSubmissionPolicy}) for more details.
        \item While we encourage the release of code and data, we understand that this might not be possible, so “No” is an acceptable answer. Papers cannot be rejected simply for not including code, unless this is central to the contribution (e.g., for a new open-source benchmark).
        \item The instructions should contain the exact command and environment needed to run to reproduce the results. See the NeurIPS code and data submission guidelines (\url{https://nips.cc/public/guides/CodeSubmissionPolicy}) for more details.
        \item The authors should provide instructions on data access and preparation, including how to access the raw data, preprocessed data, intermediate data, and generated data, etc.
        \item The authors should provide scripts to reproduce all experimental results for the new proposed method and baselines. If only a subset of experiments are reproducible, they should state which ones are omitted from the script and why.
        \item At submission time, to preserve anonymity, the authors should release anonymized versions (if applicable).
        \item Providing as much information as possible in supplemental material (appended to the paper) is recommended, but including URLs to data and code is permitted.
    \end{itemize}

\item {\bf Experimental Setting/Details}
    \item[] Question: Does the paper specify all the training and test details (e.g., data splits, hyperparameters, how they were chosen, type of optimizer, etc.) necessary to understand the results?
    \item[] Answer: \answerYes{} 
    \item[] Justification: We give experimental settings in Sec. \ref{exp_setup}, which is further detailed in Appendix \ref{append. exp}. Implementation details can be found in the anonymous link at \url{https://anonymous.4open.science/r/MSG-16E9}.
    \item[] Guidelines:
    \begin{itemize}
        \item The answer NA means that the paper does not include experiments.
        \item The experimental setting should be presented in the core of the paper to a level of detail that is necessary to appreciate the results and make sense of them.
        \item The full details can be provided either with the code, in appendix, or as supplemental material.
    \end{itemize}

\item {\bf Experiment Statistical Significance}
    \item[] Question: Does the paper report error bars suitably and correctly defined or other appropriate information about the statistical significance of the experiments?
    \item[] Answer: \answerYes{} 
    \item[] Justification: 
    In the experiment, we perform $10$ independent runs for each case, and report the mean with standard derivations in Sec. \ref{exp}.
    \item[] Guidelines:
    \begin{itemize}
        \item The answer NA means that the paper does not include experiments.
        \item The authors should answer "Yes" if the results are accompanied by error bars, confidence intervals, or statistical significance tests, at least for the experiments that support the main claims of the paper.
        \item The factors of variability that the error bars are capturing should be clearly stated (for example, train/test split, initialization, random drawing of some parameter, or overall run with given experimental conditions).
        \item The method for calculating the error bars should be explained (closed form formula, call to a library function, bootstrap, etc.)
        \item The assumptions made should be given (e.g., Normally distributed errors).
        \item It should be clear whether the error bar is the standard deviation or the standard error of the mean.
        \item It is OK to report 1-sigma error bars, but one should state it. The authors should preferably report a 2-sigma error bar than state that they have a 96\% CI, if the hypothesis of Normality of errors is not verified.
        \item For asymmetric distributions, the authors should be careful not to show in tables or figures symmetric error bars that would yield results that are out of range (e.g. negative error rates).
        \item If error bars are reported in tables or plots, The authors should explain in the text how they were calculated and reference the corresponding figures or tables in the text.
    \end{itemize}

\item {\bf Experiments Compute Resources}
    \item[] Question: For each experiment, does the paper provide sufficient information on the computer resources (type of compute workers, memory, time of execution) needed to reproduce the experiments?
    \item[] Answer: \answerYes{} 
    \item[] Justification: In Sec. \ref{exp_setup}, we provide the information on the computer resources: NVIDIA GeForce RTX 4090 GPU 24GB memory, and AMD EPYC 9654 CPU with 96-Core Processor.
    \item[] Guidelines:
    \begin{itemize}
        \item The answer NA means that the paper does not include experiments.
        \item The paper should indicate the type of compute workers CPU or GPU, internal cluster, or cloud provider, including relevant memory and storage.
        \item The paper should provide the amount of compute required for each of the individual experimental runs as well as estimate the total compute. 
        \item The paper should disclose whether the full research project required more compute than the experiments reported in the paper (e.g., preliminary or failed experiments that didn't make it into the paper). 
    \end{itemize}
    
\item {\bf Code Of Ethics}
    \item[] Question: Does the research conducted in the paper conform, in every respect, with the NeurIPS Code of Ethics \url{https://neurips.cc/public/EthicsGuidelines}?
    \item[] Answer: \answerYes{} 
    \item[] Justification: The research conducted in the paper conforms with the NeurIPS Code of Ethics.
    \item[] Guidelines:
    \begin{itemize}
        \item The answer NA means that the authors have not reviewed the NeurIPS Code of Ethics.
        \item If the authors answer No, they should explain the special circumstances that require a deviation from the Code of Ethics.
        \item The authors should make sure to preserve anonymity (e.g., if there is a special consideration due to laws or regulations in their jurisdiction).
    \end{itemize}

\item {\bf Broader Impacts}
    \item[] Question: Does the paper discuss both potential positive societal impacts and negative societal impacts of the work performed?
    \item[] Answer: \answerYes{} 
    \item[] Justification: We discuss both potential positive societal impacts and negative societal impacts of the work performed in Sec. \ref{impact}. 
    \item[] Guidelines:
    \begin{itemize}
        \item The answer NA means that there is no societal impact of the work performed.
        \item If the authors answer NA or No, they should explain why their work has no societal impact or why the paper does not address societal impact.
        \item Examples of negative societal impacts include potential malicious or unintended uses (e.g., disinformation, generating fake profiles, surveillance), fairness considerations (e.g., deployment of technologies that could make decisions that unfairly impact specific groups), privacy considerations, and security considerations.
        \item The conference expects that many papers will be foundational research and not tied to particular applications, let alone deployments. However, if there is a direct path to any negative applications, the authors should point it out. For example, it is legitimate to point out that an improvement in the quality of generative models could be used to generate deepfakes for disinformation. On the other hand, it is not needed to point out that a generic algorithm for optimizing neural networks could enable people to train models that generate Deepfakes faster.
        \item The authors should consider possible harms that could arise when the technology is being used as intended and functioning correctly, harms that could arise when the technology is being used as intended but gives incorrect results, and harms following from (intentional or unintentional) misuse of the technology.
        \item If there are negative societal impacts, the authors could also discuss possible mitigation strategies (e.g., gated release of models, providing defenses in addition to attacks, mechanisms for monitoring misuse, mechanisms to monitor how a system learns from feedback over time, improving the efficiency and accessibility of ML).
    \end{itemize}
    
\item {\bf Safeguards}
    \item[] Question: Does the paper describe safeguards that have been put in place for responsible release of data or models that have a high risk for misuse (e.g., pretrained language models, image generators, or scraped datasets)?
    \item[] Answer: \answerNA{} 
    \item[] Justification: The paper poses no such risks.
    \item[] Guidelines:
    \begin{itemize}
        \item The answer NA means that the paper poses no such risks.
        \item Released models that have a high risk for misuse or dual-use should be released with necessary safeguards to allow for controlled use of the model, for example by requiring that users adhere to usage guidelines or restrictions to access the model or implementing safety filters. 
        \item Datasets that have been scraped from the Internet could pose safety risks. The authors should describe how they avoided releasing unsafe images.
        \item We recognize that providing effective safeguards is challenging, and many papers do not require this, but we encourage authors to take this into account and make a best faith effort.
    \end{itemize}

\item {\bf Licenses for existing assets}
    \item[] Question: Are the creators or original owners of assets (e.g., code, data, models), used in the paper, properly credited and are the license and terms of use explicitly mentioned and properly respected?
    \item[] Answer: \answerYes{} 
    \item[] Justification: We properly cite the public datasets and open source Python library (e.g., \texttt{Geoopt} and \texttt{SpikingJelly})  in Sec. \ref{exp_setup} and Appendix \ref{append. data}.
    \item[] Guidelines:
    \begin{itemize}
        \item The answer NA means that the paper does not use existing assets.
        \item The authors should cite the original paper that produced the code package or dataset.
        \item The authors should state which version of the asset is used and, if possible, include a URL.
        \item The name of the license (e.g., CC-BY 4.0) should be included for each asset.
        \item For scraped data from a particular source (e.g., website), the copyright and terms of service of that source should be provided.
        \item If assets are released, the license, copyright information, and terms of use in the package should be provided. For popular datasets, \url{paperswithcode.com/datasets} has curated licenses for some datasets. Their licensing guide can help determine the license of a dataset.
        \item For existing datasets that are re-packaged, both the original license and the license of the derived asset (if it has changed) should be provided.
        \item If this information is not available online, the authors are encouraged to reach out to the asset's creators.
    \end{itemize}

\item {\bf New Assets}
    \item[] Question: Are new assets introduced in the paper well documented and is the documentation provided alongside the assets?
    \item[] Answer: \answerYes{} 
    \item[] Justification: We give the source code of the proposed model with documentation at \url{https://anonymous.4open.science/r/MSG-16E9}.
    \item[] Guidelines:
    \begin{itemize}
        \item The answer NA means that the paper does not release new assets.
        \item Researchers should communicate the details of the dataset/code/model as part of their submissions via structured templates. This includes details about training, license, limitations, etc. 
        \item The paper should discuss whether and how consent was obtained from people whose asset is used.
        \item At submission time, remember to anonymize your assets (if applicable). You can either create an anonymized URL or include an anonymized zip file.
    \end{itemize}

\item {\bf Crowdsourcing and Research with Human Subjects}
    \item[] Question: For crowdsourcing experiments and research with human subjects, does the paper include the full text of instructions given to participants and screenshots, if applicable, as well as details about compensation (if any)? 
    \item[] Answer: \answerNA{} 
    \item[] Justification: The paper does not involve crowdsourcing nor research with human subjects.
    \item[] Guidelines:
    \begin{itemize}
        \item The answer NA means that the paper does not involve crowdsourcing nor research with human subjects.
        \item Including this information in the supplemental material is fine, but if the main contribution of the paper involves human subjects, then as much detail as possible should be included in the main paper. 
        \item According to the NeurIPS Code of Ethics, workers involved in data collection, curation, or other labor should be paid at least the minimum wage in the country of the data collector. 
    \end{itemize}

\item {\bf Institutional Review Board (IRB) Approvals or Equivalent for Research with Human Subjects}
    \item[] Question: Does the paper describe potential risks incurred by study participants, whether such risks were disclosed to the subjects, and whether Institutional Review Board (IRB) approvals (or an equivalent approval/review based on the requirements of your country or institution) were obtained?
    \item[] Answer: \answerNA{} 
    \item[] Justification: The paper does not involve crowdsourcing nor research with human subjects.
    \item[] Guidelines:
    \begin{itemize}
        \item The answer NA means that the paper does not involve crowdsourcing nor research with human subjects.
        \item Depending on the country in which research is conducted, IRB approval (or equivalent) may be required for any human subjects research. If you obtained IRB approval, you should clearly state this in the paper. 
        \item We recognize that the procedures for this may vary significantly between institutions and locations, and we expect authors to adhere to the NeurIPS Code of Ethics and the guidelines for their institution. 
        \item For initial submissions, do not include any information that would break anonymity (if applicable), such as the institution conducting the review.
    \end{itemize}

\end{enumerate}

\end{document}